\documentclass[11pt]{article}

\usepackage[preprint]{acl}

\usepackage{times}
\usepackage{latexsym}

\usepackage[T1]{fontenc}

\usepackage[utf8]{inputenc}

\usepackage{microtype}

\usepackage{inconsolata}

\usepackage{graphicx}
\usepackage{tabularx}

\usepackage{amsmath}
\usepackage{amssymb}
\usepackage{booktabs}
\usepackage{pifont}
\usepackage{array}
\newcommand{\cmark}{\ding{51}}
\newcommand{\xmark}{\ding{55}}

\title{Can Agents Relive Past Experiences? Diagnosing Event-Affect Decoupling in Affective-Episodic Retrospection}

\author{
  Tingna Xie\textsuperscript{1} \quad
  Ming Wang\textsuperscript{1} \quad
  Shi Feng\textsuperscript{1,\textdagger} \quad
  Yongkang Liu\textsuperscript{2} \\
  \textsuperscript{1}Northeastern University \\
  \textsuperscript{2}Northeastern University at Qinhuangdao \\
  \textsuperscript{\textdagger}Corresponding author: fengshi@cse.neu.edu.cn
}

\begin{document}
\maketitle
\begin{abstract}

As memory-augmented agents increasingly support long-term and personalized interactions, reliable recall of emotional experiences while preserving their associated affect becomes necessary.  However, existing evaluations usually assess emotion accuracy and factual memory grounding separately, overlooking event-affect correspondence. In this Problem, we formulate Affective-Episodic Retrospection (AER) as a task that explicitly
evaluates the recovery of past events, associated affects, and their
correspondence, and define Event-Affect Decoupling as a failure mode where the
correct emotion is recalled but associated with the wrong episode.To diagnose Event-Affect Decoupling, we propose EM-TRACE, a stage-wise framework that decomposes AER into retrieval, affect recovery, and realization. On a controlled GoEmotions-based benchmark, Emotion-only retrieval achieves
28.65\% affect recovery but retrieves the target episode for only 4.44\% of
queries, yielding a UAR of 24.21\%.
Further analysis reveals that event retrieval and affect assignment introduce
distinct binding failures.
These findings show that reliable AER requires
grounded recall of emotional experiences, where episode retrieval, affect
recovery, event-affect binding, and generation quality should be evaluated
jointly. Our code is available at  \url{https://anonymous.4open.science/r/A412/}.

\end{abstract}

\begin{figure}[ht]
  \centering
  \includegraphics[width=\columnwidth]{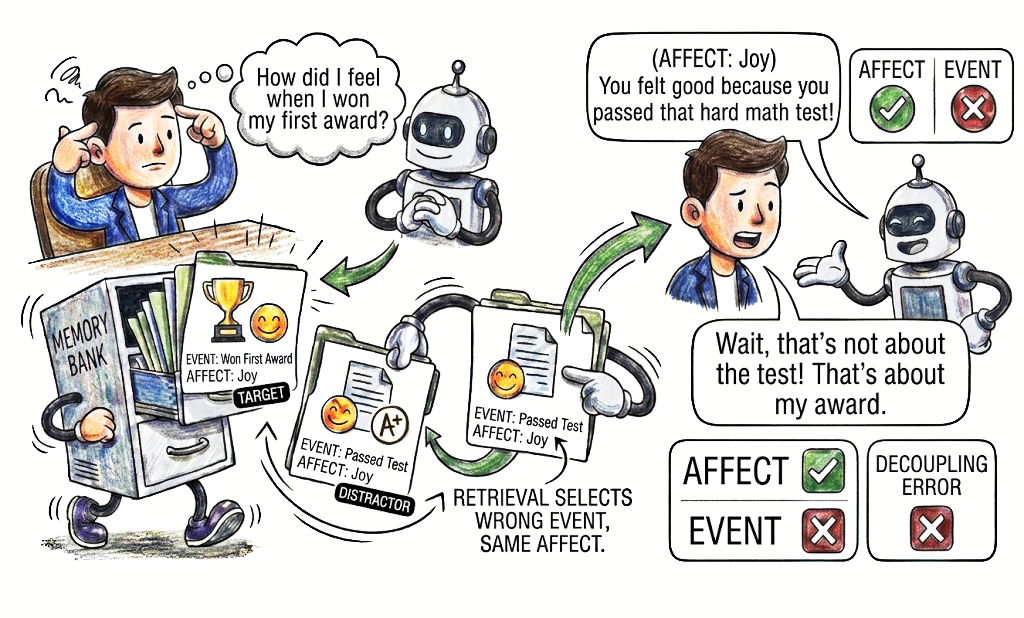}
  \caption{An emotion-correct response can rest on the wrong stored experience.}
  \label{fig:intro}
\end{figure}

\section{Introduction}

As memory-augmented language models increasingly support persistent and
personalized interactions, assistants are expected to recall information and
experiences beyond the current context window, extending memory capabilities
toward affective retrospection of past experiences
\citep{lewis2020retrieval,packer2023memgpt,zhong2024memorybank,xu2025amem}. Human episodic memory research suggests that similar affective states can be
associated with different experiences
\citep{conway2009episodic,yonelinas2015slow}. Therefore, for emotional experiences, reliable recall should require not only
recovering the relevant event and affect, but also preserving the correspondence between them. We formalize this task as Affective-Episodic Retrospection (AER).

Existing memory benchmarks primarily evaluate factual or event recall and
reasoning over stored information
\citep{maharana-etal-2024-evaluating,wu2025longmemeval}.
More recent work further incorporates user states, constraints, and emotional
needs into memory retrieval
\citep{li-etal-2026-locomo,fu-etal-2026-enpmr}.
These extensions focus on selecting memories relevant to the user's current
state or emotional needs. On the response side, emotion-oriented dialogue and faithfulness evaluations
separately assess affective responses and factual support
\citep{rashkin-etal-2019-towards,liu-etal-2021-towards,
min-etal-2023-factscore,es-etal-2024-ragas}.
Taken together, these evaluations do not directly test whether the recovered
affect remains grounded in the correct episode.

Figure~\ref{fig:intro} illustrates this blind spot.
A user asks how they felt when they won their first award.
The memory bank contains this experience with a joy label, but also another
joy-related experience, such as passing a test.
If retrieval selects the second memory, the assistant may still generate a
response expressing joy, and an emotion classifier may judge the response as
correct. Yet the response is grounded in the wrong episode: the affect is correct, but
it is bound to the wrong event.
We term this distinct failure mode Event-Affect Decoupling: correct
affect recovery despite incorrect episodic grounding. To systematically investigate this failure mode, we formulate three research questions about AER.

\textbf{RQ1: Does emotion correctness reliably indicate episodic grounding?}
We examine how often correct emotion labels coincide with retrieval of the
target episode.

\textbf{RQ2: Where do Event-Affect Binding failures originate?}
Such failures may arise from incorrect episode retrieval or from affect
assignment after successful retrieval, but existing evaluations do not
separate these sources.

\textbf{RQ3: How do emotion control and upstream recall errors affect
generated responses?}
Although emotion control can improve affective expression, it remains unclear
whether such improvements preserve factual support or how upstream retrieval
and affect assignment errors influence the final response.

To systematically analyze Event-Affect Decoupling, we introduce EM-TRACE, a stage-wise
diagnostic framework for AER.
EM-TRACE decomposes the recall process into three stages:
Episode Retrieval, Affect Recovery and Assignment, and Emotion Realization.
This decomposition enables stage-level attribution of failures arising from
episode retrieval, affect assignment, and final generation.
It further formalizes Event-Affect Binding as an evaluation dimension, where
Binding@1 measures jointly correct event-affect recovery and Ungrounded Affect
Rate (UAR) captures correct affect attached to incorrect episodes.

We study Event-Affect Decoupling in a controlled episodic recall setting
constructed from GoEmotions
\citep{demszky-etal-2020-goemotions}.
We compare retrieval strategies and emotion realization mechanisms to analyze
event-affect binding and the grounding of generated responses.
Oracle interventions further disentangle the effects of episode retrieval
errors and affect assignment errors on affect realization and factual
grounding.
The results reveal three main findings.
First, affect correctness alone is insufficient to indicate episodic grounding,
as a system may recover the correct affect while retrieving an unrelated
episode.
Second, binding failures arise from both episode retrieval and subsequent affect
assignment.
Third, emotion control enhances affect realization, but improved emotional
expression does not necessarily preserve factual support for the recalled
episode.




Our contributions are:
\begin{itemize}
  \setlength{\itemsep}{1pt}
  \setlength{\parskip}{1pt}
  \setlength{\parsep}{1pt}
  \item  We identify and define Event-Affect Decoupling, where the recovered affect
is correct but associated with the wrong episode.
  We formalize Affective-Episodic Retrospection (AER) as recovering a stored
  event, its associated affect, and their correspondence.
  
  \item We introduce EM-TRACE, a stage-wise diagnostic framework that formalizes
  Event-Affect Binding with Binding@1 and UAR and localizes failures across
  episode retrieval, affect assignment, and emotion realization.
  
  \item We construct a controlled benchmark and systematically evaluate retrieval,
  affect control, and oracle interventions, revealing that affect correctness can
  overestimate episodic grounding and that event and affect errors have distinct
  effects.
\end{itemize}

\begin{figure*}[t]
  \centering
  \includegraphics[width=\textwidth]{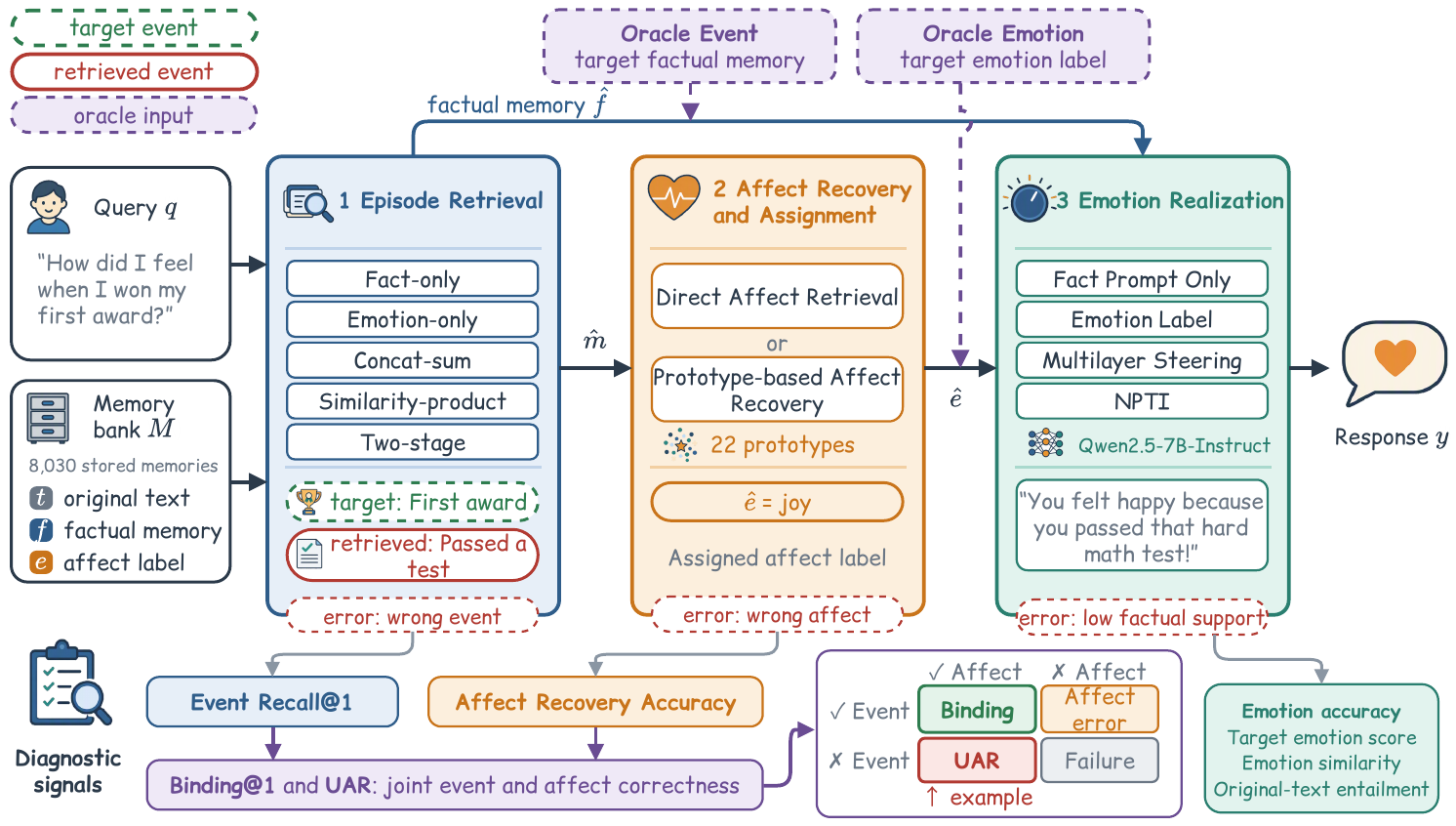}
  \caption{EM-TRACE decomposes episodic emotion recall into Episode Retrieval,
  Affect Recovery and Assignment, and Emotion Realization, with oracle
  interventions and diagnostic signals for stage-wise analysis.}
  \label{fig:framework}
\end{figure*}

\section{Related Work}

\paragraph{Episodic and Affective Memory in LLM Agents}
Long-term memory systems support persistent interactions by storing and
retrieving information beyond the context window
\citep{packer2023memgpt,zhong2024memorybank,park2023generative,xu2025amem,rasmussen2025zep,chhikara2025mem0}.
LoCoMo and LongMemEval evaluate conversational recall and reasoning across
sessions \citep{maharana-etal-2024-evaluating,wu2025longmemeval}.
Episodic memory benchmarks further assess recall of specific events and
reasoning over their temporal and spatial context \citep{DBLP:conf/iclr/HuetB025}.
Recent work incorporates affective information into memory management
\citep{lu2025damm} and selects memories according to users' emotional needs
\citep{fu-etal-2026-enpmr}.
MemEmo evaluates emotional information extraction, memory updating, and
question answering \citep{liu2026mememo}.
These developments make affect recovery an important requirement for
recalling emotional experiences.
However, different episodes may share the same affect, so correct affect
recovery alone cannot establish whether the target episode has been recalled.


\paragraph{Affective Responses and Factual Grounding}
Empathetic dialogue and emotional support benchmarks evaluate how well
responses acknowledge users' feelings and provide emotional support
\citep{rashkin-etal-2019-towards,liu-etal-2021-towards}.
Factuality and faithfulness metrics, including FActScore, RAGAS, and ARES,
assess whether generated content is supported by knowledge sources or
retrieved evidence
\citep{min-etal-2023-factscore,es-etal-2024-ragas,saad-falcon-etal-2024-ares}.
These approaches assess emotional response quality and factual support,
but do not directly test whether recovered affect is grounded in the
target episode.
Therefore, our work examines whether the recalled event supports the recovered affect
and whether that event matches the target episode.

\section{Task Formulation and EM-TRACE}
\label{sec:framework}

\subsection{Controlled Episodic Emotion Recall}

In Affective-Episodic Retrospection (AER), an agent recalls a past event and
its associated affect while preserving their correspondence.
The memory bank $\mathcal{M}=\{m_j\}_{j=1}^{N}$ contains episodes
$m_j=(t_j,f_j,e_j)$, where $t_j$ is the original text,
$f_j$ is its factual representation, and $e_j$ is its emotion label.
An episode denotes the complete stored experience, while an event denotes
the factual content represented by $f_j$.
Each query $q_i$ is generated from the original text of a target episode
$m_i^*=(t_i^*,f_i^*,e_i^*)$.
The query asks the agent to recall this experience from the memory bank. Given $q$ and $\mathcal M$, the AER task requires recovering the target
event, its associated emotion, and their correspondence.
We represent this recovery target as
\begin{equation}
z^*(q,\mathcal M)=\langle f^*,e^*,r^*\rangle,
\label{eq:aer-target}
\end{equation}
where $f^*$ is the target event, $e^*$ is its emotion label, and
$r^*$ specifies the correspondence between $f^*$ and $e^*$
within the target episode.
For $q=q_i$, the target event and emotion label are $f_i^*$ and $e_i^*$.
Different episodes may share the same emotion label, so recovering that label
alone does not identify the target experience.
The correspondence represented by $r^*$ ties the recovered emotion to
the specific event being recalled.

The agent generates a response $y$ from the query and memory bank.
Let $\hat f$, $\hat e$, and $\hat r$ denote the event, emotion label, and
correspondence expressed in that response.
The predicate $A$ assesses whether the expressed emotion label matches
$e^*$.
The predicate $C$ assesses whether the expressed event matches $f^*$
and whether the emotion is linked to that event in accordance with $r^*$.
Successful AER requires both conditions,
\begingroup
\setbox0=\hbox{ $\displaystyle \operatorname{Success}_{\mathrm{AER}}(y)=A(\hat e,e^*)\land C(\hat f,\hat e,\hat r;z^*).$}
\begin{equation}
\ifdim\wd0>\dimexpr\columnwidth-3em\relax
  \resizebox{\dimexpr\columnwidth-3em\relax}{!}{\copy0}
\else
  \copy0
\fi
\label{eq:aer-success}
\end{equation}
\endgroup Event-Affect Decoupling (EAD) occurs when the recovered emotion label is
correct but its correspondence with the target event fails.
We define this failure as
\begin{equation}
\operatorname{EAD}(y)=A(\hat e,e^*)\land\neg C(\hat f,\hat e,\hat r;z^*).
\label{eq:ead-definition}
\end{equation}
In our controlled setting, we diagnose event-affect binding at the recall stage using the identity of the retrieved episode and its assigned emotion label.

\subsection{EM-TRACE }


EM-TRACE is a diagnostic framework that decomposes episodic emotion recall into
three stages (Figure~\ref{fig:framework}). Episode Retrieval selects an experience from the memory bank.
Affect Recovery and Assignment determines the affect associated with the
retrieved experience.
Emotion Realization generates a response conditioned on the recovered episode
and affect. This decomposition enables stage-wise analysis of three error sources:
incorrect episode retrieval, incorrect affect assignment, and emotion
realization without correct episodic grounding.
The diagnostic interventions used to isolate these sources are introduced in
Section~\ref{sec:realization}.

\paragraph{Episode Retrieval}

Episode Retrieval selects the candidate memory $\hat m_i$ for query $q_i$.
We define $S_F(q,j)$ and $S_E(q,j)$ as the cosine similarities between the
query and memory $j$ in factual and affective representation spaces,
respectively. Different retrieval strategies instantiate Episode Retrieval with different
combinations of these signals.
We compare five strategies:
Fact-only, Emotion-only, Concat-sum, Similarity-product, and Two-stage. Fact-only and Emotion-only rely solely on $S_F$ and $S_E$, respectively.
Concat-sum and Similarity-product fuse the two signals through addition and
multiplication, using $S_F(q,j)+S_E(q,j)$ and
$S_F(q,j)S_E(q,j)$ as their ranking scores.
Two-stage retrieves the episode using $S_F$ and performs affect assignment on the
selected memory using emotion prototypes. 

All strategies share the same memory bank, query set, encoders, similarity
functions, and Top-1 selection rule.
Prototype construction details are provided in Appendix~\ref{app:prototype}.

\paragraph{Affect Recovery and Assignment}

This stage assigns an affect label $\hat e_i$ to the retrieved memory
$\hat m_i$.
The symbol $\hat e_i$ denotes the affect assigned at this stage rather than an
emotion prediction from the generated response. For Fact-only, Emotion-only, Concat-sum, and Similarity-product, $\hat e_i$ is
the affect label stored with the retrieved memory.
Two-stage instead performs an additional affect assignment step by comparing
the retrieved memory's affect cue representation $\hat c_i$ with class-level
prototypes $p_{e,k}$:

\begin{equation}
  \hat e_i=\arg\max_{e}\,\max_{1\le k\le K}\,
  \hat c_i^{\top}p_{e,k}.
\end{equation} Each of the 11 affect categories uses $K=2$ prototypes.
Thus, Two-stage separates episode retrieval from affect assignment.


\paragraph{Emotion Realization}

Given a retrieved factual memory and an affect signal, Emotion Realization
generates the final response $y_i$.
This stage examines how upstream retrieval and affect assignment errors
influence affect realization and factual support in the generated response. We consider three affect-control mechanisms.
Normal Emotion Label provides the assigned affect as an explicit textual label.
Multilayer Steering injects an affect-specific steering direction into selected
hidden layers during generation \citep{turner2023steering,rimsky-etal-2024-steering}.
Our emotion-control adaptation of NPTI uses neuron-level activation
intervention \citep{deng2025npti}.
Detailed construction and intervention settings are provided in
Appendix~\ref{app:implementation}.

\begin{table}[t]
\centering
\small
\begin{tabular}{lll}
\hline
\textbf{Condition} & \textbf{Affect source} & \textbf{Control}\\
\hline
Fact Prompt Only & none & none\\
Normal Emotion Label & assigned & text label\\
Normal Multilayer Steering & assigned & activation\\
Normal NPTI & assigned & activation\\
\hline
\end{tabular}
\caption{Normal retrieval-conditioned generation settings.}
\label{tab:main-conditions}
\end{table}

\begin{table}[t]
\centering
\small
\begin{tabular}{lll}
\hline
\textbf{Condition} & \textbf{Event source} & \textbf{Affect source}\\
\hline
Normal & retrieved & assigned\\
Oracle Emotion & retrieved & target\\
Oracle Event & target & assigned\\
Full Oracle & target & target\\
\hline
\end{tabular}
\caption{Oracle decomposition by event and affect sources.}
\label{tab:oracle-def}
\end{table}

\subsection{Diagnostic Design}
\label{sec:realization}
\paragraph{Main Generation Conditions}
For each retrieval strategy, we evaluate retrieval-conditioned generation
under four normal settings, summarized in
Table~\ref{tab:main-conditions}. Fact Prompt Only generates responses using the retrieved factual memory without
affect control.
The remaining conditions introduce affect through explicit labels or internal
activation interventions.

\paragraph{Oracle Analysis}
To disentangle the effects of episode retrieval and affect assignment, we adopt
a $2\times2$ oracle design over event and affect sources, as shown in
Table~\ref{tab:oracle-def}. Each oracle configuration is instantiated with Emotion Label, Multilayer
Steering, and NPTI.
Oracle Emotion replaces the assigned affect with the target affect while
keeping the retrieved factual memory unchanged.
Oracle Event replaces the retrieved factual memory with the target factual
memory while keeping the assigned affect unchanged.
Full Oracle provides both the target factual memory and target affect.

\subsection{Event-Affect Binding Metrics}

We measure event-affect binding in AER by checking the retrieved episode
and its assigned emotion label before response generation.
These metrics evaluate the recalled experience, while $A$ and $C$
assess the generated response.
For each query $q_i$, let $E_i$ indicate whether the retrieved episode
matches the target episode:
\begin{equation}
E_i=\mathbb{I}[\mathrm{id}(\hat m_i)=\mathrm{id}(m_i^*)].
\label{eq:event_correct}
\end{equation}
We apply the emotion-correctness predicate $A$ to the assigned label to
measure emotion assignment:
\begin{equation}
A_i=\mathbb{I}[A(\hat e_i,e_i^*)].
\label{eq:affect_correct}
\end{equation}
For categorical labels, $A_i=1$ exactly when $\hat e_i=e_i^*$.
The assigned label $\hat e_i$ is distinct from $\hat e$, the label expressed
in the response.
Event Recall@1 and Affect Recovery Accuracy are the averages of $E_i$ and
$A_i$ over all queries.

Binding@1 measures how often the target episode and its emotion label are
both recovered correctly:
\begin{equation}
\mathrm{Binding@1}=\frac{1}{N}\sum_{i=1}^{N}E_iA_i.
\end{equation}
Ungrounded Affect Rate (UAR) measures how often a correct emotion label is
assigned to a non-target episode:
\begin{equation}
\mathrm{UAR}=\frac{1}{N}\sum_{i=1}^{N}(1-E_i)A_i.
\end{equation}
It provides a recall-stage diagnosis of EAD.
Both metrics use all queries as the denominator, so Affect Recovery
Accuracy equals Binding@1 plus UAR.
Table~\ref{tab:quadrants} summarizes the four combinations of episode and
emotion correctness.

When a method uses the retrieved episode's stored emotion label, correct
retrieval also guarantees correct emotion assignment.
Binding@1 therefore equals Event Recall@1 by construction.
Two-stage retrieval assigns emotion separately, allowing assignment errors
even after correct retrieval.
Its Binding@1 can therefore fall below Event Recall@1.

\begin{table}[t]
  \centering
  \small
  \begin{tabular}{cc l}
    \hline
    \textbf{Event} & \textbf{Affect} & \textbf{Interpretation} \\
    \hline
    \cmark & \cmark & Correct Event-Affect Binding \\
    \cmark & \xmark & Affect assignment error \\
    \xmark & \cmark & Ungrounded affect (UAR cell) \\
    \xmark & \xmark & Full recall failure \\
    \hline
  \end{tabular}
  \caption{Quadrants of event and affect correctness.}
  \label{tab:quadrants}
\end{table}




\


\begin{table*}[t]
  \centering
  \small
  \begin{tabularx}{\textwidth}{
    l
    >{\centering\arraybackslash}X
    >{\centering\arraybackslash}X
    >{\centering\arraybackslash}X
    >{\centering\arraybackslash}X
  }
    \hline
    \textbf{Method} &
    \textbf{Event Recall@1} &
    \textbf{Affect Recovery Acc.} &
    \textbf{Binding@1} &
    \textbf{UAR} \\
    \hline
    Fact-only &
    71.62 [68.90, 74.25] &
    81.32 [78.97, 83.59] &
    71.62 [68.90, 74.25] &
    9.70 [7.98, 11.51] \\

    Emotion-only &
    4.44 [3.26, 5.71] &
    28.65 [26.02, 31.37] &
    4.44 [3.26, 5.71] &
    24.21 [21.76, 26.75] \\

    Concat-sum &
    72.17 [69.45, 74.80] &
    82.41 [80.15, 84.68] &
    72.17 [69.45, 74.80] &
    10.24 [8.43, 12.06] \\

    Similarity-product &
    72.08 [69.45, 74.71] &
    82.14 [79.87, 84.41] &
    72.08 [69.45, 74.71] &
    10.06 [8.34, 11.88] \\

    Two-stage &
    71.62 [68.90, 74.25] &
    63.74 [60.83, 66.55] &
    53.58 [50.68, 56.48] &
    10.15 [8.43, 11.97] \\
    \hline
  \end{tabularx}

  \caption{Main retrieval and binding metrics across five strategies, reported as percentages over 1,103 test queries with 95\% bootstrap confidence intervals.}
  \label{tab:binding}
\end{table*}


\begin{figure*}[t]
  \centering
  \includegraphics[width=\textwidth,height=0.28\textheight,keepaspectratio]{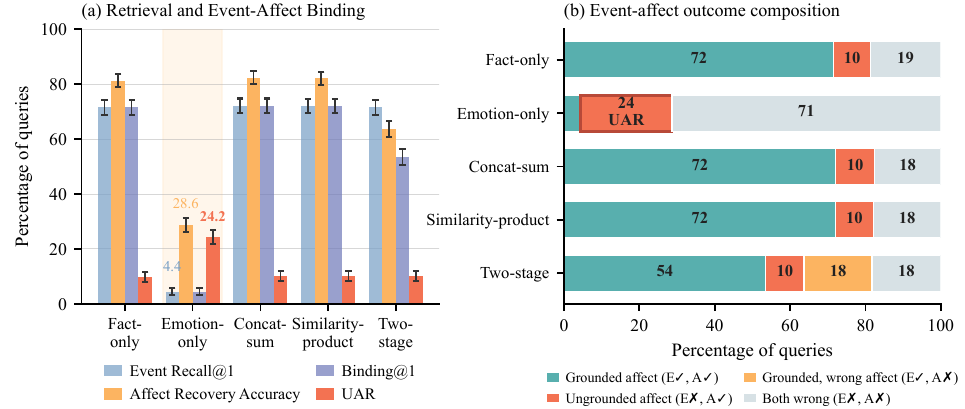}
  \caption{Retrieval and Event-Affect Binding results across five strategies. Panel (a) shows metric estimates with 95\% bootstrap confidence intervals, and panel (b) shows the four event-affect outcome proportions.}
  \label{fig:binding}
\end{figure*}

\section{Experiments}

\subsection{Experimental Setup}

\paragraph{Dataset and Memory Construction}
We construct a controlled episodic recall setting based on GoEmotions and focus
on 11 emotion categories:
admiration, amusement, curiosity, joy, confusion, sadness, disappointment,
surprise, excitement, desire, and embarrassment.
The 11 categories were selected through training-only intervention exploration
and were fixed before final test evaluation. We remove multi-label instances and extract factual memories by removing
explicit affective expressions while preserving core event content.
After factual-memory extraction and ModernBERT-based consistency filtering,
we obtain 6,927 training memories. The memory bank contains 6,927 training memories and 1,103 test memories,
resulting in 8,030 stored experiences.
For each test memory, GPT-5.6-Terra generates one evaluation query from the
original text.
All experiments use the same 1,103 queries. Test memories are included as retrieval candidates because the task evaluates
recall of stored experiences rather than generalization to unseen memories.
Steering vectors and NPTI neuron sets are estimated only from training data.
Full-memory and Train-only prototype variants are further compared as a
sensitivity analysis.


\paragraph{Evaluation Metrics}

We evaluate retrieval, emotion assignment, and their binding using the
recall-stage metrics defined in Section~\ref{sec:framework}.

For generation evaluation, emotion correctness is measured using a
ModernBERT classifier fine-tuned on GoEmotions
\citep{warner-etal-2025-smarter}.
Emotion similarity is computed as the cosine similarity between the
28-dimensional emotion distributions of the generated response and the
original text. To evaluate factual support, we use a DeBERTa-v3-large NLI model
\citep{he2023debertav3}.
The generated response is treated as the premise and the original text as the
hypothesis; the entailment probability is reported as
original-text entailment.

\paragraph{Implementation and Statistical Protocol}

We use Qwen3-Embedding-0.6B
\citep{zhang2025qwen3embedding} for factual representations,
GoEmotions-fine-tuned ModernBERT for affect representations, and
Qwen2.5-7B-Instruct
\citep{yang2024qwen25} for response generation. All generations use greedy decoding with
max\_new\_tokens=96. 

We perform query-level statistical analysis using paired bootstrap resampling
over query instances \citep{efron1993introduction,berg-kirkpatrick-etal-2012-empirical}. Main generation metrics use 10,000 resamples, while oracle analysis uses 5,000
resamples. Binary comparisons use exact McNemar tests \citep{mcnemar1947note}
with Holm correction over 77 hypotheses \citep{holm1979simple}.
All tests are two-sided \citep{dror-etal-2018-hitchhikers}. Additional implementation details are provided in
Appendix~\ref{app:implementation}.

\subsection{RQ1: Correct Affect Recovery Does Not Guarantee Episode Retrieval}


Table~\ref{tab:binding} and Figure~\ref{fig:binding} show that affect
correctness alone does not guarantee episodic grounding.
The gap is most evident in Emotion-only retrieval which achieves 28.65\% Affect
Recovery Accuracy but retrieves the target episode for only 4.44\% of queries,
leading to a UAR of 24.21\% (267/1,103).
Thus, many queries receive the correct affect from non-target episodes,
showing that emotion correctness alone can substantially overestimate
episodic recall. Factual retrieval reduces but does not eliminate this issue.
Fact-only achieves 71.62\% Event Recall@1, yet 107 of the 313 incorrectly
retrieved episodes still receive the correct affect label (UAR 9.70\%).
This shows that affect correctness can also hide episode retrieval failures. Beyond retrieval errors, Two-stage reveals that event-affect binding can fail
after successful episode retrieval.
Although its Event Recall@1 matches Fact-only, prototype-based affect
assignment reduces Affect Recovery Accuracy to 63.74\% and Binding@1 to
53.58\%, corresponding to 199 queries with correct episodes but incorrect
affect assignments.
This result indicates that episode retrieval and affect assignment introduce
distinct sources of event-affect binding failures.



\begin{figure}[t]
  \centering
  \includegraphics[width=0.80\columnwidth]{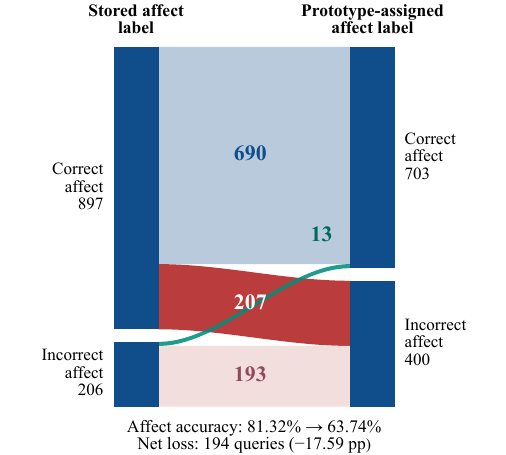}
  \caption{Affect-label correctness transitions from stored labels to prototype-assigned labels in Two-stage recall. Flow labels indicate the number of queries.}
  \label{fig:transition}
\end{figure}

\subsection{RQ2: Localizing Event-Affect Binding Failures}

\paragraph{Affect Assignment Can Break Correct Retrieval.}
Two-stage separates episode retrieval from affect assignment, allowing us to track
label transitions after retrieval (Figure~\ref{fig:transition}).
The retrieved memory carries the correct stored affect for 897 queries and an
incorrect affect for 206 queries.
The prototype assignment preserves 690 correct labels, but changes 207 correct
labels into incorrect ones, resulting in a corruption rate of 23.08\%.
It repairs only 13 incorrect cases, corresponding to a correction rate of
6.31\%.
Thus, under this setting, affect assignment introduces more incorrect
transitions than corrections, showing that successful episode retrieval does not
guarantee correct event-affect binding.



\paragraph{Prototype Source Does Not Alter the Main Pattern.}

We further examine whether prototype construction affects the observed binding
failures.
Full-memory uses all stored memories, whereas Train-only uses only the 6,927
training memories.
Changing the prototype source affects only 31 of 1,103 queries.
Affect Recovery Accuracy changes from 63.74\% to 64.19\%, and Binding@1 changes
from 53.58\% to 53.76\%.
The corruption rate remains similar (23.08\% vs. 22.52\%), while the correction
rate remains unchanged (6.31\%).
The observed pattern of binding failures remains consistent across prototype
sources.


\paragraph{Oracle Analysis Reveals Distinct Event and Affect Bottlenecks.}

\begin{figure*}[t]
  \centering
  \includegraphics[width=\textwidth]{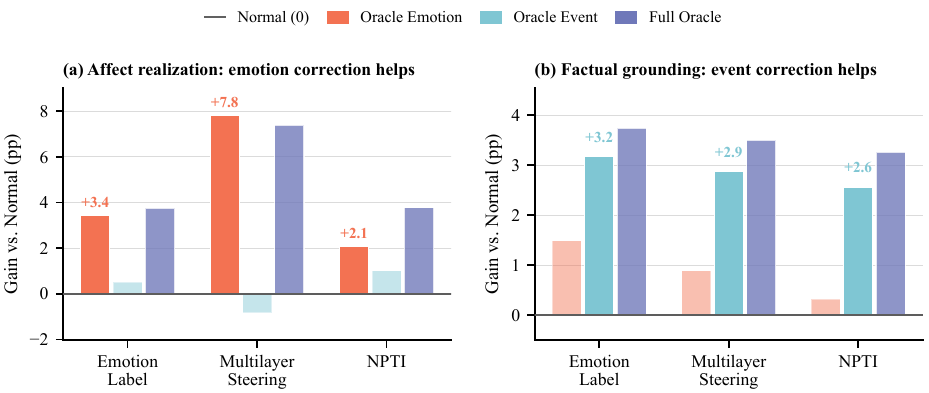}
  \caption{Mean gains over the Normal setting for different oracle interventions. Results are reported as absolute changes in percentage points (pp).}
  \label{fig:oracle}
\end{figure*}

Oracle analysis separates the effects of episode retrieval and affect assignment
errors on downstream generation.
Table~\ref{tab:oracle} reports averages over the five retrieval methods, and
Figure~\ref{fig:oracle} shows changes relative to Normal. Correcting the affect source consistently improves emotion realization.
Oracle Emotion increases emotion accuracy by 3.43, 7.80, and 2.07 points for
Emotion Label, Multilayer Steering, and NPTI, respectively.
In contrast, Oracle Event changes emotion accuracy by only +0.51, $-$0.83,
and +1.02 points.

\begin{table*}[t]
  \centering
  \small
  \setlength{\tabcolsep}{4pt}
  \begin{tabular}{lllcccc}
    \hline
    \textbf{Family} & \textbf{Condition} & \textbf{Event / Emotion} & \textbf{Emotion Acc.} & \textbf{Target Score} & \textbf{Emotion Sim.} & \textbf{Entailment} \\
    \hline
    Emotion Label & Normal & retrieved / assigned & 29.43 & 0.3098 & 0.5058 & 0.3819 \\
     & Oracle Emotion & retrieved / target & 32.86 & 0.3425 & 0.5220 & 0.3968 \\
     & Oracle Event & target / assigned & 29.94 & 0.3117 & 0.5131 & 0.4136 \\
     & Full Oracle & target / target & 33.18 & 0.3412 & 0.5261 & 0.4193 \\
    \hline
    Multilayer Steering & Normal & retrieved / assigned & 40.13 & 0.3791 & 0.5326 & 0.3485 \\
     & Oracle Emotion & retrieved / target & 47.92 & 0.4426 & 0.5690 & 0.3574 \\
     & Oracle Event & target / assigned & 39.29 & 0.3791 & 0.5350 & 0.3772 \\
     & Full Oracle & target / target & 47.51 & 0.4442 & 0.5681 & 0.3835 \\
    \hline
    NPTI & Normal & retrieved / assigned & 28.78 & 0.2969 & 0.5159 & 0.3717 \\
     & Oracle Emotion & retrieved / target & 30.84 & 0.3233 & 0.5290 & 0.3749 \\
     & Oracle Event & target / assigned & 29.79 & 0.3054 & 0.5242 & 0.3973 \\
     & Full Oracle & target / target & 32.55 & 0.3318 & 0.5393 & 0.4043 \\
    \hline
  \end{tabular}
  \caption{Means over five retrieval methods, with emotion accuracy in percent. Changes in the text use unrounded means.}
  \label{tab:oracle}
\end{table*}

The paired comparisons confirm this difference:
Oracle Emotion improves emotion accuracy in 15 of 15 method-family cells,
with 11 comparisons remaining significant after Holm correction, whereas no
Oracle Event comparison on emotion accuracy remains significant.
The largest gains occur for Multilayer Steering with Emotion-only retrieval
(+19.67 points, 95\% CI [17.14, 22.21]) and Two-stage retrieval
(+8.07 points, 95\% CI [6.26, 9.88]).

Event correction mainly affects factual support.
Oracle Event improves original-text entailment by 0.032, 0.029, and 0.026 on
average for the three realization families, with the largest gains appearing
for Emotion-only retrieval (0.071-0.098).
Together, these results reveal distinct bottlenecks: affect errors mainly limit
emotion realization, whereas event errors mainly reduce factual support.

\subsection{RQ3: Emotion Control Improves Expression Without Restoring
Episodic Grounding}

Table~\ref{tab:gen} summarizes generation performance averaged over the five
retrieval methods.
Across all three realization mechanisms, introducing affect control improves
emotion realization over Fact Prompt Only.
Emotion Label increases emotion accuracy from 23.21\% to 29.43\% under Normal
and 32.86\% with Oracle Emotion.
Multilayer Steering achieves the largest gains, reaching 40.13\% under Normal
and 47.92\% with Oracle Emotion, while NPTI provides more modest improvements
(28.78\% and 30.84\%, respectively).

\begin{table}[t]
  \centering
  \footnotesize
  \setlength{\tabcolsep}{2.5pt}
  \begin{tabular}{lcccc}
    \hline
    \textbf{Condition} & \textbf{Acc.} & \textbf{Tgt.} & \textbf{Sim.} & \textbf{Ent.} \\
    \hline
    Fact Prompt Only & 23.21 & 0.2382 & 0.4845 & 0.3526 \\
    Normal Emotion Label & 29.43 & 0.3098 & 0.5058 & 0.3819 \\
    Oracle Emotion Label & 32.86 & 0.3425 & 0.5220 & 0.3968 \\
    Normal Multilayer Steering & 40.13 & 0.3791 & 0.5326 & 0.3485 \\
    Oracle Multilayer Steering & 47.92 & 0.4426 & 0.5690 & 0.3574 \\
    Normal NPTI & 28.78 & 0.2969 & 0.5159 & 0.3717 \\
    Oracle NPTI & 30.84 & 0.3233 & 0.5290 & 0.3749 \\
    \hline
  \end{tabular}
  \caption{Means over five retrieval methods. Acc.\ is emotion accuracy in
  percent, Tgt.\ the target emotion score, Sim.\ the emotion similarity, and
  Ent.\ the original-text entailment.}
  \label{tab:gen}
\end{table}

However, stronger affect realization does not consistently translate into
better factual support.
Normal Multilayer Steering improves emotion accuracy over Normal Emotion Label
(40.13\% vs.\ 29.43\%), but yields lower entailment (0.3485 vs.\ 0.3819).
The same trend holds under Oracle Emotion: Multilayer Steering reaches 47.92\%
emotion accuracy while remaining 0.0394 below Oracle Emotion Label in entailment.
This gap is clearest under poor retrieval.
With Oracle Multilayer Steering, Emotion-only retrieval achieves the highest
emotion accuracy (50.05\%) but the lowest entailment (0.3066), whereas other
retrieval strategies remain between 0.3694 and 0.3716.
Thus, stronger emotional expression alone does not restore episodic grounding.

\section{Conclusion}

In this work, we identify and define Event-Affect Decoupling, where an agent recovers the
correct affect but associates it with the wrong episode.
We formulate Affective-Episodic Retrospection (AER) as the task of recovering
a past  event, its associated affect, and their correspondence.
Through EM-TRACE, our experiments reveal binding failures in both episode
retrieval and subsequent affect assignment.
Oracle interventions show that affect errors primarily limit emotional
expression, while event errors primarily reduce factual support.
Stronger emotion control improves expression but does not reliably restore
episodic grounding.
These findings highlight event-affect correspondence as a central requirement for reliable recall of emotional experiences.
Future evaluation should therefore move beyond affect correctness alone and
examine whether recovered affects remain bound to their corresponding
experiences.

\section*{Limitations}

\paragraph{Controlled episodic recall setting.}
EM-TRACE evaluates episodic emotion recall in a controlled benchmark constructed
from GoEmotions rather than naturally occurring user interactions.
The memory bank consists of pre-built records and does not capture factors
such as memory formation, updating, or evolving personal context over time.
Therefore, our findings characterize event-affect retrieval and binding under
a controlled protocol, while further validation on naturally collected
interaction data is needed.

\paragraph{Emotion representation.}
EM-TRACE represents affect as discrete emotion categories and performs affect
recovery and binding analysis based on these labels.
However, human emotional experiences are often continuous, mixed, and
dependent on contextual factors.
Therefore, the current benchmark evaluates event-affect binding under a
categorical emotion setting, and future work could investigate richer affect
representations, such as continuous emotion spaces or context-aware affective
states.

\paragraph{Binding evaluation scope.}
Binding@1 and Ungrounded Affect Rate (UAR) define event-affect binding based on
record identity and affect labels, enabling controlled diagnosis of different
failure sources.
However, this formulation provides a discrete measurement of binding and may
not fully capture semantic consistency between recalled experiences and
generated responses.
Future work should explore more fine-grained human evaluations of event
identity, affect expression, and memory grounding.

\section*{Ethical Considerations}

The data derive from GoEmotions, a public corpus of Reddit comments with emotion annotations, and every stored memory in our bank is a simulated record.
The generated texts describe no real personal experience and should not be read as such.
Memory-augmented assistants that store emotional experiences raise privacy questions, since stored memories can reveal sensitive information about a user, and emotion control could be used to steer users or to imitate empathy.
Our diagnostics are intended to help detect responses that sound emotionally appropriate while resting on the wrong memory, which supports safer deployment of such assistants.
Automatic emotion labels may misrepresent how people from different backgrounds express emotion, and the categorical label space is a coarse description of affect.

\section*{AI Usage Statement}

During the preparation of this manuscript, we used AI assistants for several
supporting tasks, including data preparation assistance, evaluation query
construction, and manuscript editing.

For data preparation, Xiaomi MiMo-Pro 2.5 was used to assist with extracting
factual memory representations from the original texts. For evaluation
construction, GPT-5.6-Terra was used to assist in generating evaluation queries
from the original texts of the test set. These AI-generated outputs were used
as preliminary candidates and were subsequently checked and verified by the
authors before being included in the experimental pipeline.

For manuscript preparation, large language models were used to assist with
English writing refinement, improving clarity and readability, and adjusting
LaTeX formatting to comply with the ARR submission template. The use of AI
assistants was limited to language refinement and presentation support rather
than the development of research ideas, experimental methodology, or scientific
claims.

All AI-assisted outputs were manually reviewed and verified by the authors.
The authors independently designed the research questions, methodology,
experimental settings, analyses, and conclusions, and reviewed all
AI-modified text and formatting changes before submission.

\bibliography{custom}

\appendix

\section{Data Construction Details}
\label{app:data}

\subsection{Category selection and split roles}

Category selection was performed before final test evaluation.  Starting from
the complete 28-label GoEmotions inventory, we conducted exploratory
intervention experiments for each emotion separately using training data.  For
each label, candidate injection layers were searched and the affect-injection
behavior at its best-performing layer or layers was examined.  The 11 labels
carried forward in this study were those identified through this training-only
exploration.  Neither the final test questions nor their generated responses
were used to select the category set.

Table~\ref{tab:data-flow} summarizes the distinct roles of the training and
test splits.  Both splits are restricted to single-label records belonging to
the selected 11 categories.  The training records additionally undergo the
dual-description consistency check because they are used to construct the
affect-intervention artifacts.  Test records do not contribute to steering
vectors or NPTI neuron sets and therefore do not undergo this check.  They do,
however, undergo factual-memory extraction because their factual memories are
stored as retrieval candidates and targets.  All 1,103 eligible test records
are retained; no additional test-record filtering is applied.

\begin{table*}[t]
  \centering
  \footnotesize
  \setlength{\tabcolsep}{5pt}
  \begin{tabular}{p{2.6cm}p{5.0cm}p{5.6cm}r}
    \hline
    \textbf{Split / stage} & \textbf{Inclusion or operation} & \textbf{Purpose} & \textbf{Retained} \\
    \hline
    Category selection
      & Training-only exploration of all 28 labels, including a search for effective injection layers per label
      & Select the 11 study emotions before final test evaluation
      & 11 labels \\
    Training candidates
      & Single-label records in the selected categories with a non-empty extracted factual memory
      & Candidate data for intervention-artifact construction
      & 9,754 \\
    Accepted training memories
      & Training candidates passing the affective- and neutral-description consistency checks
      & Construct steering vectors and NPTI neuron sets; populate the memory bank
      & 6,927 \\
    Eligible test records
      & Single-label records in the selected categories; factual-memory extraction is applied, with no additional record filtering
      & Populate the memory bank and provide retrieval targets and evaluation queries; excluded from steering-vector and NPTI-set construction
      & 1,103 \\
    Final memory bank
      & Accepted training memories plus all eligible test memories
      & Retrieval candidate bank
      & 8,030 \\
    \hline
  \end{tabular}
  \caption{Data flow and split-specific processing.}
  \label{tab:data-flow}
\end{table*}

\subsection{Construction models and decoding settings}

Table~\ref{tab:construction-generation} reports the generation settings that
can affect the constructed data.  Factual-memory extraction for both the
training and test splits uses the recorded API model alias
\texttt{mimo-v2.5-pro}.  For API-hosted models, the original runs did not
persist provider-side snapshot dates or revision hashes; consequently, these
stages are reproducible at the model-alias and request-parameter level, but not
as byte-identical provider snapshots.  The locally loaded
Qwen2.5-7B-Instruct checkpoint has artifact revision
\texttt{077a6aff06a7d75066ecc3abe12ce0d673873603}.

\begin{table*}[t]
  \centering
  \footnotesize
  \setlength{\tabcolsep}{3pt}
  \begin{tabularx}{\textwidth}{
    p{2.1cm}
    p{2.8cm}
    p{0.7cm}
    p{0.8cm}
    p{1.6cm}
    p{1.3cm}
    X
  }
    \hline
    \textbf{Stage} & \textbf{Model} & \textbf{Temp.} &
    \textbf{Top-$p$} & \textbf{Max output} &
    \textbf{Retries} & \textbf{Failure handling} \\
    \hline

    Factual-memory extraction
      & mimo-v2.5-pro
      & 0.0
      & not sent
      & API default
      & 5 API attempts
      & Invalid batches fall back to per-item parsing; unresolved items are marked error/null \\

    Initial affective description
      & Qwen2.5-7B-Instruct
      & 0.0
      & n/a
      & 96 tokens
      & none
      & Short outputs fail; classifier mismatches enter the retry pipeline \\

    Initial neutral description
      & Qwen2.5-7B-Instruct
      & 0.0
      & n/a
      & 96 tokens
      & none
      & Generation or neutral-QC failures do not enter the accepted artifact \\

    Affective-description retry
      & Qwen2.5-7B-Instruct
      & 0.0
      & n/a
      & 128 tokens
      & 3 prompts
      & Each output is reclassified; all failures are recorded as errors \\

    Test-question generation
      & gpt-5.6-terra
      & not sent
      & not sent
      & 100 tokens
      & 3 API attempts
      & Errors are checkpointed; complete coverage is required for final output \\

    \hline
  \end{tabularx}

  \caption{Model aliases and decoding settings for data construction.
  A temperature of 0.0 invokes deterministic greedy decoding for the local
  Qwen model. ``Not sent'' means that the corresponding API-provider default
  was used.}
  \label{tab:construction-generation}
\end{table*}

\subsection{Filtering and factual-memory extraction}

Multi-label records are removed before factual-memory construction.
Among the remaining records in the 11 selected categories, 9,754 training examples yield a valid factual memory.
Factual-memory extraction uses mimo-v2.5-pro. The system instruction is:
\begin{quote}
\small
\textbf{System:} Extract objective factual memory from text. Remove emotion, subjective judgment, and affective social wording. Keep only facts stated in the text; do not add or translate. Preserve perspective.\\[2pt]
\textbf{Examples:} ``Thank you friend'' $\rightarrow$ ``Express gratitude to a friend.''; ``My neighbor has been playing loud music since 2 AM, I am absolutely furious and can't sleep!'' $\rightarrow$ ``My neighbor has been playing music since 2 AM. I can't sleep.''
\end{quote}
For batched extraction, the user message supplies a JSON list of \texttt{item\_id}-text pairs and requests
\texttt{\{``items'': [\{``item\_id'': integer, ``factual\_memory'': string|null\}, ...]\}}.
If batch parsing fails, the implementation falls back to the single-record user template
\texttt{Text: \{original\_text\}} followed by
\texttt{JSON: \{``factual\_memory'': ...\}}.
The request sets \texttt{temperature=0.0}, does not send \texttt{top\_p}, and
does not explicitly set \texttt{max\_tokens} or
\texttt{max\_completion\_tokens}; the output limit is therefore the API
default.  API requests permit up to five attempts.  Invalid structured batch
responses, including missing or mismatched item identifiers, trigger
single-record fallback with the same prompt.  Items that remain unsuccessful
are marked with an extraction error and a null factual memory and can be
reprocessed by the retry script using the same prompt and settings.  Successful
items are checkpointed.  Transport-level failures abort the affected batch
rather than fabricating a result.  A record enters the 9,754-candidate pool only
when extraction returns a non-empty factual memory.

\subsection{Dual description consistency check}

For each factual-memory candidate, Qwen2.5-7B-Instruct produces an affective description and a neutral description. The following templates define the initial affective rendering and the neutral rendering.

\paragraph{Affective-description prompt.}
\begin{quote}
\small
\textbf{System:} You write a short emotional description from an objective fact.\\[2pt]
\textbf{Rules:} Express the target emotion through concrete wording and
perspective.  Do not explain the task or directly repeat the target emotion
label.  Do not use template phrases such as ``I feel [emotion]'' or ``This is
[emotion].''  Do not add new entities or background.  Preserve the language of
the fact.  Output only the emotional description text.\\[2pt]
\textbf{User:} Create an emotional description.\\
\texttt{Objective fact: \{factual\_memory\}}\\
\texttt{Target emotion: \{emotion\_label\}}\\
Write naturally without naming the target emotion label directly.  Avoid
template phrases like ``I feel [emotion].''\\
\texttt{Emotional description:}
\end{quote}

\paragraph{Neutral-description prompt.}
\begin{quote}
\small
\textbf{System:} You write a short objective description from a fact.\\[2pt]
\textbf{Rules:} Keep the wording natural and emotionally neutral.  Do not
explain the task or add new entities or background.  Preserve the language of
the fact.  Output only the description text.\\[2pt]
\textbf{User:} Create an objective description.\\
\texttt{Objective fact: \{factual\_memory\}}\\
Keep the description emotionally neutral.\\
\texttt{Objective description:}
\end{quote}

Both initial descriptions use deterministic greedy decoding with
temperature=0.0, at most 96 new tokens, and a minimum accepted length
of three tokens.  The client seed is 42, although no sampling occurs at zero
temperature.

Both descriptions are checked with the 28-way ModernBERT-GoEmotions classifier.
The affective description must receive the target emotion as its top-1 prediction, and the neutral description must receive neutral.
An affective description that fails the initial classifier gate is regenerated
with the following predefined prompts in order.  Each prompt is paired with the
same factual memory and target emotion.

\paragraph{Affective retry 1.}
\begin{quote}
\small
You are given an objective fact and a target emotion.

Rewrite the fact as a short first-person or natural-language description with a
clear emotional tone.  The target emotion should be recognizable from wording,
tone, and framing, but the emotion label itself must not appear.

\textbf{Rules:} (1) Preserve all factual information exactly.  Do not add,
remove, or alter facts, causes, motives, entities, times, or intentions.  (2)
Make the emotional stance clear enough that an external emotion classifier
would identify the target emotion.  (3) Do not state the target emotion label
directly, and avoid obvious variants of that label.  (4) Keep the same language
as the original fact.  (5) Output only the emotional description.
\end{quote}

\paragraph{Affective retry 2.}
\begin{quote}
\small
You are given an objective fact and a target emotion.

Express the same fact with stronger affective framing while staying
fact-faithful.  Use natural phrasing that makes the target emotion salient
through perspective and emphasis, without naming the emotion label.

\textbf{Rules:} (1) Do not invent background, causes, motives, relationships,
or consequences.  (2) Do not delete or weaken the core fact.  (3) Do not output
analysis, labels, explanations, quotation marks, the target emotion label, or
obvious variants of that label.  (4) Keep the same language as the original
fact.  (5) Output one concise emotional description.
\end{quote}

\paragraph{Affective retry 3.}
\begin{quote}
\small
You are given an objective fact and a target emotion.

Write a concise emotional rendering of the fact.  The wording should make the
target emotion obvious from context and framing.

\textbf{Rules:} (1) Preserve the factual meaning and all stated entities.  (2)
Do not introduce new events, intentions, causes, or outcomes.  (3) Do not
explicitly mention the target emotion label, and avoid obvious variants of that
label.  (4) Avoid generic templates such as ``I feel ...'' or ``This is ....''
(5) Keep the same language as the original fact.  (6) Output only the emotional
description.
\end{quote}

Retry generation uses greedy decoding with at most 128 new tokens.  After every
attempt, the output is reclassified and checked for direct use of the emotion
label.  The first passing output is selected in the order original, retry 1,
retry 2, and retry 3.  If none passes, the record is marked as a retry error and
does not enter the accepted training artifact.
The finalized bank retains the classifier-accepted affective description and the previously gated neutral description for each record.
Only records passing both checks are retained:
\begin{equation}
\begin{aligned}
9{,}754\ &\text{factual-memory candidates}\\
&\longrightarrow 6{,}927\ \text{training memories}.
\end{aligned}
\end{equation}

\subsection{Test-question construction}

For each of the 1,103 eligible test records, GPT-5.6-Terra generates one question from the record's \emph{original text}, not from the 6,927-record training memory set.
The system instruction is ``You create concise evaluation questions for memory retrieval systems.''
The user template asks for one natural question that (i) contains sufficient factual and contextual clues, (ii) creates a realistic setting in which the answer can preserve the source memory, (iii) preserves speaker perspective and interaction direction, (iv) introduces no new event, background, intention, motivation, or request, (v) does not state an emotion label, and (vi) does not copy the original wording.
After the fixed in-context examples, the final request is
\begin{quote}
\small
\texttt{Now generate a query for this memory:}\\[2pt]
\texttt{Memory:}\\
\texttt{\{original\_text\}}\\[2pt]
\texttt{Output only the query.}
\end{quote}
The API request sends neither temperature nor top-$p$, and therefore uses the
provider defaults for this GPT-5-style endpoint.  It sets
\texttt{max\_completion\_tokens=100} and permits up to three request attempts;
there is no separate content-level retry prompt.  Empty source text, empty
responses, and requests that remain unsuccessful are checkpointed as errors
rather than treated as valid questions.  Successful results are reused on
subsequent runs, and the final ordered output requires complete coverage of all
selected records.
The script initializes its client-side random state with seed 42 for record
selection and key ordering; no explicit model-side sampling seed is passed to the API.
This client-side seed is distinct from the prototype and bootstrap seeds reported below.

\section{Retrieval and Prototype Details}
\label{app:prototype}

\subsection{Retrieval representations}

Factual retrieval uses 1,024-dimensional Qwen3-Embedding-0.6B representations obtained by last-token pooling and L2 normalization.
Emotion cues use 768-dimensional masked-mean representations from layer 22 of ModernBERT-GoEmotions, followed by L2 normalization.
Layer 22 was selected through a training-only sweep over layers 1-22 using Recall@1, Recall@3, and Recall@5; no final test question was used for this selection.

\subsection{Prototype construction}

The prototype bank groups the emotion cue embeddings by affect label and builds two prototypes per label.
Within one label, spherical k-means splits the L2-normalized embeddings into
$K=2$ clusters \citep{dhillon2001concept}.
For each cluster, the Weiszfeld iteration finds the geometric median $p$
that minimizes $\sum_i\lVert p-\hat c_i\rVert_2$ \citep{weiszfeld2009point}.
The result is L2 normalized to give a prototype $\hat p_{e,k}$.
Assignment uses the maximum cosine similarity over the two prototypes of each label, which the equation in Section~\ref{sec:framework} states.
The Full-memory prototype bank uses the cue representations and affect labels of all 8,030 stored memories: 6,927 training memories and 1,103 test memories.
This is the transductive protocol analyzed in the main text.
Test records are used as stored retrieval/prototype keys, but never as intervention-training examples.
The steering vectors and NPTI neuron sets attached to each label use training data only in both protocols; no test emotion label is used to construct an intervention.
The Train-only bank uses the 6,927 training memories as members.
The main prototype configuration uses base seed $20260922$ and initializes the clustering for class index $r$ with $20260922+r$.
Spherical k-means runs for at most 100 iterations; Weiszfeld iteration runs for at most 1,000 iterations with tolerance $10^{-7}$.
The resulting prototypes are L2 normalized and matched with cosine similarity.
Stage-1 retrieval is deterministic exhaustive scoring over the fixed memory bank, with fixed candidate order and tie-breaking.

\subsection{Full-memory and Train-only sensitivity}

Compared with the Full-memory result, the Train-only assignment changes the label of 31 of 1,103 queries.
The corresponding Affect Recovery, Binding@1, quadrant counts, correction rate, corruption rate, and transition counts appear in Table~\ref{tab:sens} and Figure~\ref{fig:transition}.

\begin{table}[t]
  \centering
  \footnotesize
  \setlength{\tabcolsep}{2pt}
  \begin{tabular}{lccc}
    \hline
    \textbf{Metric} & \textbf{Full-mem.} & \textbf{Train-only} & \textbf{Change} \\
    \hline
    Event Recall@1 (\%) & 71.62 & 71.62 & 0 \\
    Affect Recovery Acc. (\%) & 63.74 & 64.19 & +0.45 \\
    Binding@1 (\%) & 53.58 & 53.76 & +0.18 \\
    Event\cmark{} Affect\cmark{} & 591 & 593 & +2 \\
    Event\cmark{} Affect\xmark{} & 199 & 197 & $-$2 \\
    Event\xmark{} Affect\cmark{} & 112 & 115 & +3 \\
    Event\xmark{} Affect\xmark{} & 201 & 198 & $-$3 \\
    Correction rate (\%) & 6.31 & 6.31 & 0 \\
    Corruption rate (\%) & 23.08 & 22.52 & $-$0.56 \\
    \hline
  \end{tabular}
  \caption{Two-stage with Full-memory and Train-only prototypes.}
  \label{tab:sens}
\end{table}

\subsection{Per-emotion retrieval and binding}
\label{app:per-emotion-binding}

Table~\ref{tab:per-emotion-binding} reports class counts and retrieval-stage
metrics within each target-emotion class.  The test split is already highly
imbalanced before either retrieval method is evaluated: admiration
contains 272 test queries, whereas embarrassment contains 21.  This
approximately 13-fold difference is therefore a property of the dataset split,
not an outcome of Fact-only or Two-stage retrieval.  We report both micro
averages, which weight classes by their test frequencies, and unweighted macro
averages over the 11 classes.  The Two-stage columns use the Full-memory
prototype configuration reported in the main results.

\begin{table*}[t]
  \centering
  \small
  \setlength{\tabcolsep}{5pt}
  \begin{tabularx}{\textwidth}{
    >{\raggedright\arraybackslash}p{2.0cm}
    >{\centering\arraybackslash}p{1.2cm}
    >{\centering\arraybackslash}p{1.0cm}
    *{8}{>{\centering\arraybackslash}X}
  }
    \hline
    & \multicolumn{2}{c}{\textbf{Count}}
    & \multicolumn{4}{c}{\textbf{Fact-only (\%)}}
    & \multicolumn{4}{c}{\textbf{Two-stage (\%)}} \\
    \cline{2-3}\cline{4-7}\cline{8-11}

    \textbf{Emotion} & \textbf{Train} & \textbf{Test}
    & \textbf{Event} & \textbf{Affect} & \textbf{Bind.} & \textbf{UAR}
    & \textbf{Event} & \textbf{Affect} & \textbf{Bind.} & \textbf{UAR} \\
    \hline

    Admiration     & 1,857 & 272 & 66.18 & 84.56 & 66.18 & 18.38 & 66.18 & 71.69 & 51.47 & 20.22 \\
    Amusement      &   996 & 149 & 72.48 & 81.21 & 72.48 &  8.72 & 72.48 & 75.17 & 65.77 &  9.40 \\
    Curiosity      &   845 & 162 & 77.16 & 85.19 & 77.16 &  8.02 & 77.16 & 69.14 & 60.49 &  8.64 \\
    Joy            &   519 &  81 & 70.37 & 71.60 & 70.37 &  1.23 & 70.37 & 49.38 & 45.68 &  3.70 \\
    Confusion      &   516 &  86 & 80.23 & 83.72 & 80.23 &  3.49 & 80.23 & 46.51 & 45.35 &  1.16 \\
    Sadness        &   536 &  85 & 63.53 & 80.00 & 63.53 & 16.47 & 63.53 & 58.82 & 44.71 & 14.12 \\
    Disappointment &   458 &  77 & 75.32 & 77.92 & 75.32 &  2.60 & 75.32 & 54.55 & 51.95 &  2.60 \\
    Surprise       &   529 &  74 & 74.32 & 75.68 & 74.32 &  1.35 & 74.32 & 48.65 & 45.95 &  2.70 \\
    Excitement     &   350 &  46 & 71.74 & 82.61 & 71.74 & 10.87 & 71.74 & 67.39 & 58.70 &  8.70 \\
    Desire         &   183 &  50 & 80.00 & 86.00 & 80.00 &  6.00 & 80.00 & 68.00 & 62.00 &  6.00 \\
    Embarrassment  &   138 &  21 & 52.38 & 61.90 & 52.38 &  9.52 & 52.38 & 52.38 & 42.86 &  9.52 \\

    \hline

    Macro average  & - & - & 71.25 & 79.13 & 71.25 & 7.88
    & 71.25 & 60.15 & 52.27 & 7.89 \\

    Micro average  & 6,927 & 1,103 & 71.62 & 81.32 & 71.62 & 9.70
    & 71.62 & 63.74 & 53.58 & 10.15 \\

    \hline
  \end{tabularx}

  \caption{Per-emotion retrieval and event-affect binding results.
  Rates are computed within each target-emotion class.
  Event denotes Event Recall@1, Affect denotes Affect Recovery Accuracy,
  Bind. denotes Binding@1, and UAR is the event-wrong/affect-correct rate.
  The macro average is the unweighted mean over the 11 emotion classes; the
  micro average is weighted by test-query count.}
  \label{tab:per-emotion-binding}
\end{table*}

Fact-only reads the stored affect label from the retrieved memory, so its
Binding@1 equals Event Recall@1 within every class.  Two-stage retains the same
episode retrieval results but its prototype assignment reduces Affect Recovery
from 81.32\% to 63.74\% micro and from 79.13\% to 60.15\% macro.  The reduction
is not confined to one dominant class: Two-stage Affect Recovery is relatively
higher for amusement (75.17\%), admiration (71.69\%), and
excitement (67.39\%), but falls below 50\% for confusion
(46.51\%), surprise (48.65\%), and joy (49.38\%).  Its lowest
Binding@1 values occur for embarrassment (42.86\%), sadness
(44.71\%), confusion (45.35\%), and joy (45.68\%).

Fact-only and Two-stage have similar UAR under both micro averaging (9.70\% vs.
10.15\%) and macro averaging (7.88\% vs. 7.89\%).  Thus, Two-stage does not
consistently convert more event-retrieval errors into affect-correct cases
across classes.  Its main aggregate loss instead comes from affect-assignment
errors on event-correct samples, reflected in the substantially lower
Binding@1.

\section{Complete Generation Results}
\label{app:gen}

We report 53 retrieval-conditioned and oracle units together with one shared Bare Query baseline.
The presentation follows the experimental structure of Section~\ref{sec:realization}: main generation conditions first, followed by Oracle Emotion, Oracle Event, and Full Oracle.
Full Oracle is independent of retrieval method because both its event and affect sources are fixed to the target; it is therefore reported once per realization mechanism.
When two conditions share the same Stage-1 event and affect input, they can yield identical generations; this accounts for identical Fact-only and Two-stage rows in several prompt-based settings.

\subsection{Main generation conditions}

Table~\ref{tab:main21} reports the 21 main-condition units: Bare Query, five Fact Prompt Only units, and the three Normal realization mechanisms under five retrieval methods each.

\begin{table*}[t]
  \centering
  \footnotesize
  \setlength{\tabcolsep}{4pt}
  \begin{tabular}{llcccc}
    \hline
    \textbf{Condition} & \textbf{Method} & \textbf{Emotion Acc. (\%) [95\% CI]} & \textbf{Target} & \textbf{Similarity} & \textbf{Entailment} \\
    \hline
    Bare Query & shared & 23.03 & 0.2409 & 0.4657 & 0.3219 \\
\hline
Fact Prompt Only & Fact-only & 23.57 [21.12, 26.11] & 0.2414 & 0.4890 & 0.3644 \\
 & Emotion-only & 22.67 [20.22, 25.20] & 0.2330 & 0.4712 & 0.3068 \\
 & Concat-sum & 23.12 [20.67, 25.66] & 0.2375 & 0.4871 & 0.3608 \\
 & Similarity-product & 23.12 [20.67, 25.66] & 0.2379 & 0.4861 & 0.3665 \\
 & Two-stage & 23.57 [21.12, 26.11] & 0.2414 & 0.4890 & 0.3644 \\
\hline
Normal Emotion Label & Fact-only & 30.92 [28.20, 33.64] & 0.3253 & 0.5085 & 0.4036 \\
 & Emotion-only & 25.02 [22.48, 27.65] & 0.2583 & 0.4763 & 0.3119 \\
 & Concat-sum & 31.10 [28.38, 33.91] & 0.3281 & 0.5150 & 0.3982 \\
 & Similarity-product & 31.01 [28.29, 33.82] & 0.3277 & 0.5126 & 0.3965 \\
 & Two-stage & 29.10 [26.38, 31.82] & 0.3098 & 0.5165 & 0.3993 \\
\hline
Normal Multilayer Steering & Fact-only & 43.79 [40.80, 46.78] & 0.4095 & 0.5427 & 0.3651 \\
 & Emotion-only & 30.37 [27.65, 33.09] & 0.2968 & 0.4759 & 0.2823 \\
 & Concat-sum & 43.61 [40.62, 46.60] & 0.4057 & 0.5445 & 0.3628 \\
 & Similarity-product & 43.61 [40.62, 46.60] & 0.4069 & 0.5447 & 0.3592 \\
 & Two-stage & 39.26 [36.36, 42.16] & 0.3768 & 0.5552 & 0.3730 \\
\hline
Normal NPTI & Fact-only & 30.10 [27.38, 32.91] & 0.3111 & 0.5224 & 0.3873 \\
 & Emotion-only & 24.48 [21.94, 27.02] & 0.2501 & 0.4767 & 0.3011 \\
 & Concat-sum & 30.19 [27.47, 32.91] & 0.3109 & 0.5262 & 0.3883 \\
 & Similarity-product & 30.10 [27.38, 32.82] & 0.3108 & 0.5266 & 0.3920 \\
 & Two-stage & 29.01 [26.38, 31.73] & 0.3016 & 0.5276 & 0.3897 \\
\hline

  \end{tabular}
  \caption{Main generation conditions. Emotion accuracy includes 95\% bootstrap intervals for retrieval-conditioned units; Bare Query is a shared point estimate.}
  \label{tab:main21}
\end{table*}

\subsection{Oracle analysis}

\subsubsection{Oracle Emotion}

Oracle Emotion keeps the normally retrieved event and replaces only the assigned affect with the target affect.
Table~\ref{tab:oracle-emotion15} reports all 15 units.

\begin{table*}[t]
  \centering
  \footnotesize
  \setlength{\tabcolsep}{4pt}
  \begin{tabular}{llcccc}
    \hline
    \textbf{Condition} & \textbf{Method} & \textbf{Emotion Acc. (\%) [95\% CI]} & \textbf{Target} & \textbf{Similarity} & \textbf{Entailment} \\
    \hline
    Oracle Emotion Label & Fact-only & 32.73 [29.92, 35.54] & 0.3446 & 0.5236 & 0.4119 \\
 & Emotion-only & 32.91 [30.10, 35.72] & 0.3299 & 0.5101 & 0.3463 \\
 & Concat-sum & 32.91 [30.10, 35.72] & 0.3467 & 0.5266 & 0.4075 \\
 & Similarity-product & 33.00 [30.19, 35.81] & 0.3467 & 0.5262 & 0.4065 \\
 & Two-stage & 32.73 [29.92, 35.54] & 0.3446 & 0.5236 & 0.4119 \\
\hline
Oracle Multilayer Steering & Fact-only & 47.33 [44.33, 50.23] & 0.4437 & 0.5666 & 0.3696 \\
 & Emotion-only & 50.05 [47.05, 53.13] & 0.4432 & 0.5762 & 0.3066 \\
 & Concat-sum & 47.60 [44.61, 50.50] & 0.4409 & 0.5682 & 0.3716 \\
 & Similarity-product & 47.33 [44.42, 50.32] & 0.4411 & 0.5674 & 0.3694 \\
 & Two-stage & 47.33 [44.33, 50.23] & 0.4437 & 0.5666 & 0.3696 \\
\hline
Oracle NPTI & Fact-only & 31.01 [28.20, 33.73] & 0.3257 & 0.5326 & 0.3912 \\
 & Emotion-only & 29.37 [26.65, 32.09] & 0.3114 & 0.5101 & 0.3041 \\
 & Concat-sum & 31.46 [28.65, 34.18] & 0.3272 & 0.5356 & 0.3933 \\
 & Similarity-product & 31.37 [28.56, 34.09] & 0.3268 & 0.5341 & 0.3947 \\
 & Two-stage & 31.01 [28.20, 33.73] & 0.3257 & 0.5326 & 0.3912 \\
\hline

  \end{tabular}
  \caption{Oracle Emotion results for the three realization mechanisms and five retrieval methods.}
  \label{tab:oracle-emotion15}
\end{table*}

\begin{figure*}[t]
  \centering
  \includegraphics[width=0.92\textwidth]{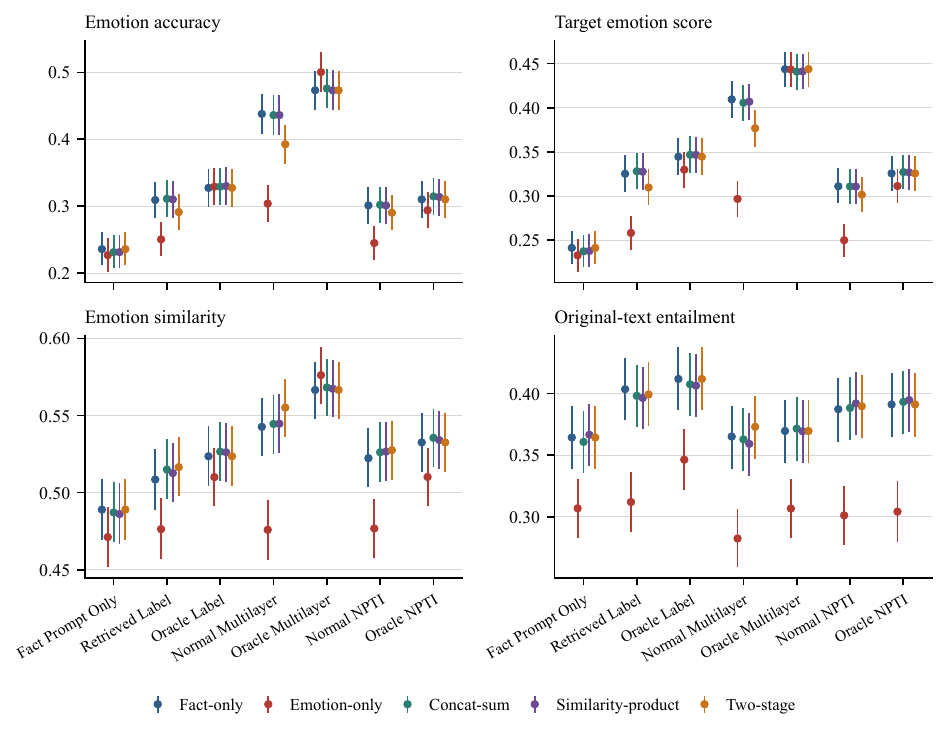}
  \caption{Normal and Oracle Emotion results across retrieval methods, with 95\% bootstrap intervals. Fact Prompt Only is included as the no-affect-control reference.}
  \label{fig:generation}
\end{figure*}

\subsubsection{Oracle Event}

Oracle Event replaces the retrieved factual memory with the target event while retaining the affect assigned by the normal retrieval pipeline.
Table~\ref{tab:oracle-event15} reports all 15 units.

\begin{table*}[t]
  \centering
  \footnotesize
  \setlength{\tabcolsep}{5pt}
  \begin{tabular}{llcccc}
    \toprule
    \textbf{Condition} & \textbf{Method} & \textbf{Emotion Acc. (\%)} & \textbf{Target} & \textbf{Similarity} & \textbf{Entailment} \\
    \midrule
    Oracle Event + Emotion Label & Fact-only & 31.19 & 0.3241 & 0.5122 & 0.4228 \\
     & Emotion-only & 25.75 & 0.2719 & 0.5015 & 0.3919 \\
     & Concat-sum & 31.64 & 0.3268 & 0.5146 & 0.4204 \\
     & Similarity-product & 31.64 & 0.3265 & 0.5145 & 0.4201 \\
     & Two-stage & 29.47 & 0.3093 & 0.5225 & 0.4128 \\
    \midrule
    Oracle Event + Multilayer Steering & Fact-only & 42.43 & 0.4076 & 0.5425 & 0.3825 \\
     & Emotion-only & 31.37 & 0.3031 & 0.4864 & 0.3537 \\
     & Concat-sum & 42.34 & 0.4042 & 0.5446 & 0.3818 \\
     & Similarity-product & 42.43 & 0.4054 & 0.5456 & 0.3825 \\
     & Two-stage & 37.90 & 0.3752 & 0.5558 & 0.3857 \\
    \midrule
    Oracle Event + NPTI & Fact-only & 30.83 & 0.3181 & 0.5271 & 0.3981 \\
     & Emotion-only & 26.38 & 0.2674 & 0.5074 & 0.3996 \\
     & Concat-sum & 31.28 & 0.3190 & 0.5301 & 0.3949 \\
     & Similarity-product & 31.19 & 0.3186 & 0.5291 & 0.3973 \\
     & Two-stage & 29.28 & 0.3042 & 0.5273 & 0.3967 \\
    \bottomrule
  \end{tabular}
  \caption{Oracle Event results for the three realization mechanisms and five retrieval methods. Values are point estimates over 1,103 test questions.}
  \label{tab:oracle-event15}
\end{table*}

\subsubsection{Full Oracle}

Full Oracle uses both the target event and target affect.
Because it is retrieval-independent, each realization mechanism contributes one shared unit rather than five duplicated rows.

\begin{table*}[t]
  \centering
  \footnotesize
  \setlength{\tabcolsep}{6pt}
  \begin{tabular*}{\textwidth}{@{\extracolsep{\fill}}llcccc}
    \toprule
    \textbf{Condition} & 
    \textbf{Method} & 
    \textbf{Emotion Acc. (\%)} & 
    \textbf{Target} & 
    \textbf{Similarity} & 
    \textbf{Entailment} \\
    \midrule

    Full Oracle + Emotion Label & shared & 33.18 & 0.3412 & 0.5261 & 0.4193 \\
    Full Oracle + Multilayer Steering & shared & 47.51 & 0.4442 & 0.5681 & 0.3835 \\
    Full Oracle + NPTI & shared & 32.55 & 0.3318 & 0.5393 & 0.4043 \\

    \bottomrule
  \end{tabular*}

  \caption{The three retrieval-independent Full Oracle units. Emotion accuracy is reported in percent.}
  \label{tab:full-oracle3}
\end{table*}

\subsection{Paired Oracle analysis}
\label{app:oracle}

For paired attribution, each retrieval-conditioned unit is compared with the shared Full Oracle output from the same realization mechanism.
This produces 60 \emph{comparison cells} ($3$ mechanisms $\times$ $5$ retrieval methods $\times$ $4$ source conditions), but not 60 distinct generations: Full Oracle is generated once per mechanism and reused only as a paired reference.
The $2\times2$ oracle grid therefore contains 48 unique units: 15 Normal, 15 Oracle Emotion, 15 Oracle Event, and 3 Full Oracle.
All paired intervals use 5,000 query-level resamples.

\begin{figure*}[t]
  \centering
  \includegraphics[width=0.85\textwidth]{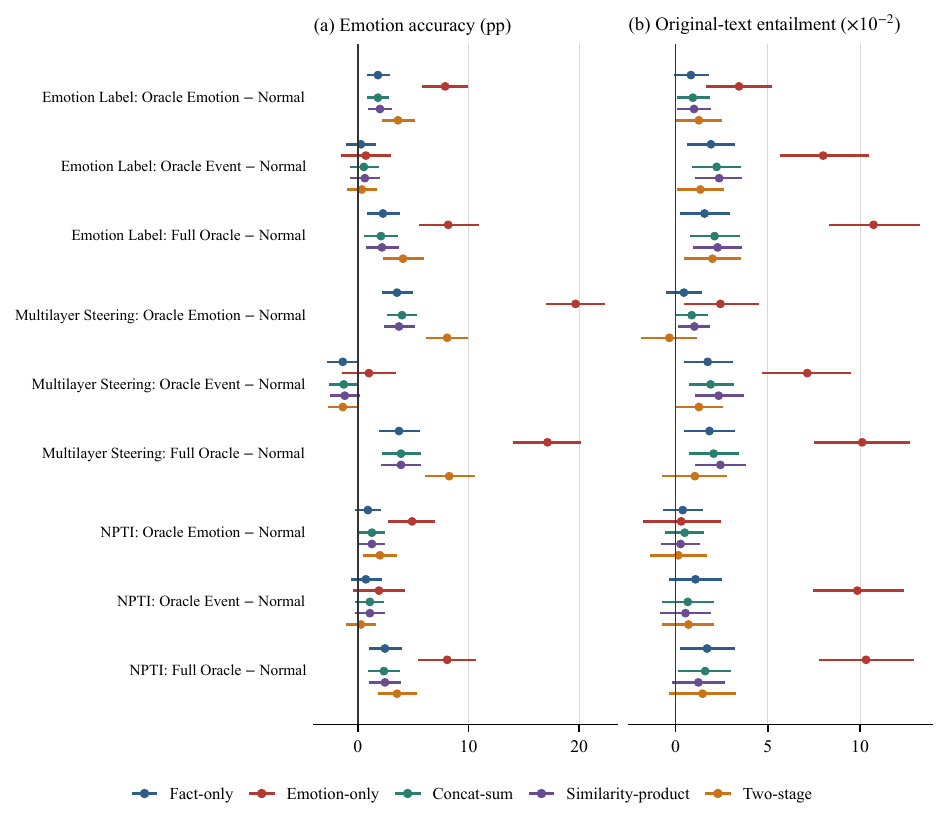}
  \caption{Paired differences to Normal with 95\% bootstrap intervals for each retrieval method and realization mechanism. The same Full Oracle generation is paired with all five retrieval-conditioned units within a mechanism, but is reported only once in the complete-results tables.}
  \label{fig:forest}
\end{figure*}

Figure~\ref{fig:forest} shows paired differences to Normal for emotion accuracy and original-text entailment.
The interaction term is $(\text{Full Oracle}-\text{Oracle Event})-(\text{Oracle Emotion}-\text{Normal})$.
Its 95\% interval includes zero for 13 of 15 method-mechanism cells in emotion accuracy, 15 of 15 in target emotion score, 14 of 15 in emotion similarity, and 15 of 15 in original-text entailment.

\begin{table}[t]
  \centering
  \small
  \begin{tabular}{lc}
    \hline
    \textbf{Contrast} & \textbf{Holm-significant} \\
    \hline
    Oracle Emotion $-$ Normal & 11 / 15 \\
    Oracle Event $-$ Normal & 0 / 15 \\
    Full Oracle $-$ Oracle Emotion & 0 / 15 \\
    Full Oracle $-$ Oracle Event & 10 / 15 \\
    Full Oracle $-$ Normal & 10 / 15 \\
    \hline
  \end{tabular}
  \caption{Holm-significant McNemar comparisons in emotion accuracy at $\alpha=0.05$.}
  \label{tab:holm}
\end{table}

Tables~\ref{tab:hyp-legacy}-\ref{tab:hyp-npti} provide the complete
machine-generated results for all 77 hypotheses.  Differences and confidence
intervals are in emotion-accuracy percentage points and follow the displayed
contrast order.  Significance is determined from the Holm-adjusted $p$ value at
$\alpha=0.05$.

\begin{table*}[t]
  \centering
  \footnotesize
  \setlength{\tabcolsep}{2pt}
  \begin{tabular*}{\textwidth}{@{\extracolsep{\fill}}lllcccc}
    \toprule
    \textbf{Mechanism} & \textbf{Metric} & \textbf{Method contrast} & \textbf{$\Delta$ [95\% CI]} & \textbf{Raw $p$} & \textbf{Holm $p$} & \textbf{Sig.} \\
    \midrule
    Retrieval & Binding@1 & two-stage $-$ fact-only & -18.04 [-20.40, -15.87] & 2.49e-60 & 1.92e-58 & Yes \\
    Retrieval & Event Recall@1 & concat-sum $-$ fact-only & 0.54 [-0.63, 1.72] & 0.451 & 1.000 & No \\
    \bottomrule
  \end{tabular*}
  \caption{The two retrieval hypotheses in the unified Holm family.  Their
  paired percentile-bootstrap intervals use 10,000 query-level resamples with
  seed 20260928.}
  \label{tab:hyp-legacy}
\end{table*}

\begin{table*}[t]
  \centering
  \footnotesize
  \setlength{\tabcolsep}{2pt}
  \begin{tabular*}{\textwidth}{@{\extracolsep{\fill}}lllcccc}
    \toprule
    \textbf{Mechanism} & \textbf{Method} & \textbf{Contrast} & \textbf{$\Delta$ [95\% CI]} & \textbf{Raw $p$} & \textbf{Holm $p$} & \textbf{Sig.} \\
    \midrule
    EL & fact-only & Oracle Emotion $-$ Normal & 1.81 [0.91, 2.81] & 3.25e-4 & 0.015 & Yes \\
    EL & fact-only & Oracle Event $-$ Normal & 0.27 [-1.00, 1.54] & 0.784 & 1.000 & No \\
    EL & fact-only & Full Oracle $-$ Oracle Emotion & 0.45 [-0.91, 1.90] & 0.609 & 1.000 & No \\
    EL & fact-only & Full Oracle $-$ Oracle Event & 1.99 [1.00, 3.08] & 1.13e-4 & 0.006 & Yes \\
    EL & fact-only & Full Oracle $-$ Normal & 2.27 [0.91, 3.72] & 0.003 & 0.105 & No \\
    EL & emotion-only & Oracle Emotion $-$ Normal & 7.89 [5.89, 9.88] & 7.70e-16 & 5.47e-14 & Yes \\
    EL & emotion-only & Oracle Event $-$ Normal & 0.73 [-1.45, 2.90] & 0.573 & 1.000 & No \\
    EL & emotion-only & Full Oracle $-$ Oracle Emotion & 0.27 [-2.27, 2.90] & 0.891 & 1.000 & No \\
    EL & emotion-only & Full Oracle $-$ Oracle Event & 7.43 [5.44, 9.52] & 7.05e-13 & 4.86e-11 & Yes \\
    EL & emotion-only & Full Oracle $-$ Normal & 8.16 [5.62, 10.79] & 5.43e-10 & 3.48e-8 & Yes \\
    EL & concat-sum & Oracle Emotion $-$ Normal & 1.81 [0.91, 2.72] & 1.80e-4 & 0.009 & Yes \\
    EL & concat-sum & Oracle Event $-$ Normal & 0.54 [-0.63, 1.81] & 0.471 & 1.000 & No \\
    EL & concat-sum & Full Oracle $-$ Oracle Emotion & 0.27 [-1.09, 1.72] & 0.798 & 1.000 & No \\
    EL & concat-sum & Full Oracle $-$ Oracle Event & 1.54 [0.63, 2.54] & 0.002 & 0.097 & No \\
    EL & concat-sum & Full Oracle $-$ Normal & 2.09 [0.63, 3.54] & 0.006 & 0.225 & No \\
    EL & similarity-product & Oracle Emotion $-$ Normal & 1.99 [1.00, 2.99] & 5.95e-5 & 0.003 & Yes \\
    EL & similarity-product & Oracle Event $-$ Normal & 0.63 [-0.63, 1.90] & 0.392 & 1.000 & No \\
    EL & similarity-product & Full Oracle $-$ Oracle Emotion & 0.18 [-1.18, 1.63] & 0.899 & 1.000 & No \\
    EL & similarity-product & Full Oracle $-$ Oracle Event & 1.54 [0.63, 2.54] & 0.002 & 0.097 & No \\
    EL & similarity-product & Full Oracle $-$ Normal & 2.18 [0.82, 3.63] & 0.004 & 0.144 & No \\
    EL & two-stage & Oracle Emotion $-$ Normal & 3.63 [2.27, 5.08] & 4.57e-7 & 2.65e-5 & Yes \\
    EL & two-stage & Oracle Event $-$ Normal & 0.36 [-0.91, 1.63] & 0.672 & 1.000 & No \\
    EL & two-stage & Full Oracle $-$ Oracle Emotion & 0.45 [-0.91, 1.90] & 0.609 & 1.000 & No \\
    EL & two-stage & Full Oracle $-$ Oracle Event & 3.72 [2.35, 5.17] & 2.79e-7 & 1.64e-5 & Yes \\
    EL & two-stage & Full Oracle $-$ Normal & 4.08 [2.36, 5.89] & 6.90e-6 & 3.80e-4 & Yes \\
    \bottomrule
  \end{tabular*}
  \caption{All 25 formal emotion-accuracy hypotheses for Emotion Label (EL)
  realization.  Intervals use 5,000 paired percentile-bootstrap resamples with
  seed 20260929.}
  \label{tab:hyp-emotion-label}
\end{table*}

\begin{table*}[t]
  \centering
  \footnotesize
  \setlength{\tabcolsep}{2pt}
  \begin{tabular*}{\textwidth}{@{\extracolsep{\fill}}lllcccc}
    \toprule
    \textbf{Mechanism} & \textbf{Method} & \textbf{Contrast} & \textbf{$\Delta$ [95\% CI]} & \textbf{Raw $p$} & \textbf{Holm $p$} & \textbf{Sig.} \\
    \midrule
    MS & fact-only & Oracle Emotion $-$ Normal & 3.54 [2.27, 4.90] & 1.52e-7 & 9.12e-6 & Yes \\
    MS & fact-only & Oracle Event $-$ Normal & -1.36 [-2.72, 0.00] & 0.063 & 1.000 & No \\
    MS & fact-only & Full Oracle $-$ Oracle Emotion & 0.18 [-1.36, 1.72] & 0.910 & 1.000 & No \\
    MS & fact-only & Full Oracle $-$ Oracle Event & 5.08 [3.63, 6.62] & 1.35e-11 & 8.92e-10 & Yes \\
    MS & fact-only & Full Oracle $-$ Normal & 3.72 [1.99, 5.53] & 3.80e-5 & 0.002 & Yes \\
    MS & emotion-only & Oracle Emotion $-$ Normal & 19.67 [17.14, 22.21] & 8.22e-49 & 6.25e-47 & Yes \\
    MS & emotion-only & Oracle Event $-$ Normal & 1.00 [-1.36, 3.35] & 0.445 & 1.000 & No \\
    MS & emotion-only & Full Oracle $-$ Oracle Emotion & -2.54 [-5.44, 0.45] & 0.103 & 1.000 & No \\
    MS & emotion-only & Full Oracle $-$ Oracle Event & 16.14 [13.60, 18.68] & 2.84e-33 & 2.13e-31 & Yes \\
    MS & emotion-only & Full Oracle $-$ Normal & 17.14 [14.14, 20.04] & 6.35e-29 & 4.70e-27 & Yes \\
    MS & concat-sum & Oracle Emotion $-$ Normal & 3.99 [2.72, 5.26] & 1.31e-10 & 8.49e-9 & Yes \\
    MS & concat-sum & Oracle Event $-$ Normal & -1.27 [-2.54, 0.00] & 0.065 & 1.000 & No \\
    MS & concat-sum & Full Oracle $-$ Oracle Emotion & -0.09 [-1.63, 1.45] & 1.000 & 1.000 & No \\
    MS & concat-sum & Full Oracle $-$ Oracle Event & 5.17 [3.72, 6.71] & 3.23e-12 & 2.20e-10 & Yes \\
    MS & concat-sum & Full Oracle $-$ Normal & 3.90 [2.27, 5.62] & 1.18e-5 & 6.39e-4 & Yes \\
    MS & similarity-product & Oracle Emotion $-$ Normal & 3.72 [2.45, 5.08] & 1.31e-8 & 8.11e-7 & Yes \\
    MS & similarity-product & Oracle Event $-$ Normal & -1.18 [-2.45, 0.09] & 0.092 & 1.000 & No \\
    MS & similarity-product & Full Oracle $-$ Oracle Emotion & 0.18 [-1.36, 1.72] & 0.906 & 1.000 & No \\
    MS & similarity-product & Full Oracle $-$ Oracle Event & 5.08 [3.63, 6.62] & 5.77e-12 & 3.86e-10 & Yes \\
    MS & similarity-product & Full Oracle $-$ Normal & 3.90 [2.18, 5.62] & 1.48e-5 & 7.83e-4 & Yes \\
    MS & two-stage & Oracle Emotion $-$ Normal & 8.07 [6.26, 9.88] & 2.69e-18 & 1.94e-16 & Yes \\
    MS & two-stage & Oracle Event $-$ Normal & -1.36 [-2.63, -0.09] & 0.053 & 1.000 & No \\
    MS & two-stage & Full Oracle $-$ Oracle Emotion & 0.18 [-1.36, 1.72] & 0.910 & 1.000 & No \\
    MS & two-stage & Full Oracle $-$ Oracle Event & 9.61 [7.71, 11.51] & 1.45e-24 & 1.06e-22 & Yes \\
    MS & two-stage & Full Oracle $-$ Normal & 8.25 [6.17, 10.43] & 8.80e-14 & 6.16e-12 & Yes \\
    \bottomrule
  \end{tabular*}
  \caption{All 25 formal emotion-accuracy hypotheses for Multilayer Steering (MS)
  realization.  Intervals use 5,000 paired percentile-bootstrap resamples with
  seed 20260929.}
  \label{tab:hyp-multilayer}
\end{table*}

\begin{table*}[t]
  \centering
  \footnotesize
  \setlength{\tabcolsep}{2pt}
  \begin{tabular*}{\textwidth}{@{\extracolsep{\fill}}lllcccc}
    \toprule
    \textbf{Mechanism} & \textbf{Method} & \textbf{Contrast} & \textbf{$\Delta$ [95\% CI]} & \textbf{Raw $p$} & \textbf{Holm $p$} & \textbf{Sig.} \\
    \midrule
    NPTI & fact-only & Oracle Emotion $-$ Normal & 0.91 [-0.18, 1.99] & 0.154 & 1.000 & No \\
    NPTI & fact-only & Oracle Event $-$ Normal & 0.73 [-0.54, 2.09] & 0.332 & 1.000 & No \\
    NPTI & fact-only & Full Oracle $-$ Oracle Emotion & 1.54 [0.09, 2.99] & 0.046 & 1.000 & No \\
    NPTI & fact-only & Full Oracle $-$ Oracle Event & 1.72 [0.73, 2.72] & 0.001 & 0.058 & No \\
    NPTI & fact-only & Full Oracle $-$ Normal & 2.45 [1.09, 3.90] & 0.001 & 0.058 & No \\
    NPTI & emotion-only & Oracle Emotion $-$ Normal & 4.90 [2.81, 6.89] & 5.82e-6 & 3.26e-4 & Yes \\
    NPTI & emotion-only & Oracle Event $-$ Normal & 1.90 [-0.36, 4.17] & 0.115 & 1.000 & No \\
    NPTI & emotion-only & Full Oracle $-$ Oracle Emotion & 3.17 [0.54, 5.71] & 0.018 & 0.628 & No \\
    NPTI & emotion-only & Full Oracle $-$ Oracle Event & 6.17 [3.99, 8.25] & 2.07e-8 & 1.26e-6 & Yes \\
    NPTI & emotion-only & Full Oracle $-$ Normal & 8.07 [5.53, 10.52] & 1.65e-9 & 1.04e-7 & Yes \\
    NPTI & concat-sum & Oracle Emotion $-$ Normal & 1.27 [0.18, 2.36] & 0.029 & 0.951 & No \\
    NPTI & concat-sum & Oracle Event $-$ Normal & 1.09 [-0.18, 2.27] & 0.119 & 1.000 & No \\
    NPTI & concat-sum & Full Oracle $-$ Oracle Emotion & 1.09 [-0.36, 2.45] & 0.162 & 1.000 & No \\
    NPTI & concat-sum & Full Oracle $-$ Oracle Event & 1.27 [0.36, 2.27] & 0.016 & 0.564 & No \\
    NPTI & concat-sum & Full Oracle $-$ Normal & 2.36 [1.00, 3.72] & 0.001 & 0.058 & No \\
    NPTI & similarity-product & Oracle Emotion $-$ Normal & 1.27 [0.18, 2.36] & 0.034 & 1.000 & No \\
    NPTI & similarity-product & Oracle Event $-$ Normal & 1.09 [-0.18, 2.36] & 0.119 & 1.000 & No \\
    NPTI & similarity-product & Full Oracle $-$ Oracle Emotion & 1.18 [-0.27, 2.54] & 0.130 & 1.000 & No \\
    NPTI & similarity-product & Full Oracle $-$ Oracle Event & 1.36 [0.36, 2.36] & 0.011 & 0.384 & No \\
    NPTI & similarity-product & Full Oracle $-$ Normal & 2.45 [1.09, 3.81] & 7.30e-4 & 0.034 & Yes \\
    NPTI & two-stage & Oracle Emotion $-$ Normal & 1.99 [0.54, 3.45] & 0.010 & 0.382 & No \\
    NPTI & two-stage & Oracle Event $-$ Normal & 0.27 [-1.00, 1.54] & 0.784 & 1.000 & No \\
    NPTI & two-stage & Full Oracle $-$ Oracle Emotion & 1.54 [0.09, 2.99] & 0.046 & 1.000 & No \\
    NPTI & two-stage & Full Oracle $-$ Oracle Event & 3.26 [1.99, 4.62] & 2.03e-6 & 1.16e-4 & Yes \\
    NPTI & two-stage & Full Oracle $-$ Normal & 3.54 [1.90, 5.26] & 7.80e-5 & 0.004 & Yes \\
    \bottomrule
  \end{tabular*}
  \caption{All 25 formal emotion-accuracy hypotheses for NPTI realization.
  Intervals use 5,000 paired percentile-bootstrap resamples with seed 20260929.}
  \label{tab:hyp-npti}
\end{table*}

\begin{table}[t]
  \centering
  \small
  \setlength{\tabcolsep}{3pt}
  \begin{tabular}{p{5.6cm}c}
    \hline
    \textbf{Audit question} & \textbf{Answer} \\
    \hline
    Memory bank has 8,030 rows (6,927 training and 1,103 test) & Yes \\
    Stage-1 input is the test-question embedding & Yes \\
    Test-memory factual vectors are Stage-1 candidates & Yes \\
    Gold factual memory is directly supplied in non-oracle prompts & No \\
    Full-memory prototypes use test emotion labels & Yes \\
    Steering/NPTI artifacts use test data or labels & No \\
    Train-only prototypes use test emotion labels & No \\
    NLI premise is the response; hypothesis is the original text & Yes \\
    Bare Query rows enter retrieval-method aggregation & No \\
    Result inventory contains 53 unique units plus one shared baseline & Yes \\
    \hline
  \end{tabular}
  \caption{Protocol audit.}
  \label{tab:audit}
\end{table}

\section{Evaluator and Statistical Details}
\label{app:evaluators}

\subsection{Emotion evaluation}

The ModernBERT-GoEmotions classification head produces sigmoid probabilities over the complete 28-label GoEmotions inventory.
Emotion accuracy compares the highest-scoring one of these 28 labels with the target label; target emotion score is the predicted probability assigned to that target.
Emotion similarity is the cosine similarity between the 28-dimensional probability vectors of the generated response and the matching record's original text.

Every target label is one of the 11 study emotions.
For readable confusion matrices only, predictions outside this set are grouped as Other: anger, annoyance, approval, caring, disapproval, disgust, fear, gratitude, grief, love, nervousness, optimism, pride, realization, relief, remorse, and neutral.
This grouping is applied only after inference; the evaluator itself remains 28-way, and every Other prediction is incorrect for emotion accuracy.

\subsection{Original-text entailment}

Original-text entailment uses the entailment probability of the DeBERTa-v3-large NLI evaluator.
The generated response is the \emph{premise}, and the original text of the matching test record is the \emph{hypothesis}.
The reported value is the softmax probability of the entailment class among entailment, neutral, and contradiction.
This directional score measures support for the reference text and is not interpreted as symmetric semantic equivalence.

\subsection{Statistical protocol}

Main results resample the 1,103 test questions 10,000 times with seed 20260928; the oracle decomposition uses 5,000 resamples with seed 20260929.
The resampling unit is \texttt{query\_row\_id}, and every resample preserves pairing across retrieval methods, generation conditions, and oracle settings.
All reported bootstrap intervals are percentile intervals: their endpoints are
the 2.5th and 97.5th percentiles of the paired bootstrap distribution.  No
basic or BCa transformation is applied.
Binary paired comparisons use exact two-sided McNemar tests.
The unified family contains 77 hypotheses: 75 oracle contrasts (five contrasts across five retrieval methods and three realization mechanisms) and two retrieval comparisons.
One Holm correction is applied over this complete family: raw $p$ values are
sorted, multiplied by the corresponding step-down factor, constrained by a
cumulative maximum, and truncated at one.  Holm-adjusted $p<0.05$ leaves 32
hypotheses significant and 45 non-significant.  The complete results appear in
Tables~\ref{tab:hyp-legacy}-\ref{tab:hyp-npti}.

\section{Generation and Intervention Implementation}
\label{app:implementation}

\subsection{Response-generation templates}

The Fact Prompt Only condition uses the following template.
\begin{quote}
\small
\texttt{Retrieved factual memory (use it as context when relevant;}\\
\texttt{do not mention this retrieval process):}\\
\texttt{\{retrieved\_factual\_memory\}}\\[2pt]
\texttt{User query:}\\
\texttt{\{evaluation\_query\}}\\[2pt]
\texttt{Generate a helpful response.}
\end{quote}

The label conditions insert the block \texttt{Emotional state:} followed by the
affect label between the memory block and the query block.
Multilayer Steering uses the Fact Prompt Only template, and the affect label
selects the corresponding steering vector.
NPTI uses the same retrieved-memory and query fields with the same
\texttt{Generate a helpful response.} suffix, while the affect label selects
the corresponding neuron set.
The shared Bare Query baseline omits the retrieved-memory block and uses
\texttt{User query: \{evaluation\_query\}} followed by
\texttt{Generate a helpful response.}
In the Oracle Event and Full Oracle conditions, the factual memory of the target
record replaces the retrieved factual memory in the template.
Within each realization mechanism, the oracle conditions do not change the test
question, mechanism-specific response template, decoding rule, or evaluator.

\subsection{Multilayer Steering}

For emotion $e$ and layer $l$, Multilayer Steering uses the mean difference
\begin{equation}
  v_e^{(l)}=\frac{1}{N_e}\sum_{i\in\mathcal{V}_e}\left(h^{(l)}_{i,e}-h^{(l)}_{i,\mathrm{neutral}}\right),
\end{equation}
where $h^{(l)}_{i,e}$ and $h^{(l)}_{i,\mathrm{neutral}}$ are the hidden states for the affective and neutral descriptions of the same factual memory.
For every selected layer, the steering-bank construction applies an
emotion-layer-specific coefficient $\alpha_e^{(l)}$,
\begin{equation}
  \widetilde v_e^{(l)}=\alpha_e^{(l)}v_e^{(l)}.
\end{equation}
The resulting pre-scaled vector $\widetilde v_e^{(l)}$ is added directly to the
last-token residual representation at transformer block $l$.  Layer indices in
Table~\ref{tab:steering-layers} are one-based.  No second global scaling factor
is applied during generation.

\begin{table}[t]
  \centering
  \footnotesize
  \setlength{\tabcolsep}{3pt}
  \begin{tabular}{lp{3.4cm}c}
    \hline
    \textbf{Emotion} & \textbf{Layer: coefficient} & \textbf{Total $\alpha$} \\
    \hline
    admiration     & 20: 0.75; 21: 0.75             & 1.50 \\
    amusement      & 20: 1.25                        & 1.25 \\
    curiosity      & 20: 1.00                        & 1.00 \\
    joy            & 20: 1.00                        & 1.00 \\
    confusion      & 22: 0.6667; 23: 0.6667; 24: 0.6667 & 2.00 \\
    sadness        & 21: 0.375; 22: 1.125            & 1.50 \\
    disappointment & 23: 2.50                        & 2.50 \\
    surprise       & 20: 0.375; 22: 1.125            & 1.50 \\
    excitement     & 19: 0.25; 20: 0.75; 22: 0.25   & 1.25 \\
    desire         & 21: 0.375; 22: 0.375; 23: 0.75 & 1.50 \\
    embarrassment  & 21: 0.6375; 22: 1.0625         & 1.70 \\
    \hline
  \end{tabular}
  \caption{Emotion-specific residual-stream layers and steering coefficients.
  Total $\alpha$ is the sum of the layer-specific coefficients and is reported
  only as a configuration summary; inference injects each pre-scaled layer
  vector separately.}
  \label{tab:steering-layers}
\end{table}

Candidate residual layers and intervention strengths were explored before the
main evaluation through single-layer scans, multilayer combinations,
layer-weight-ratio sweeps, and total-strength sweeps.  The explored strengths
included 0.1, 0.25, 0.5, 1.0, 1.25, 1.5, and 2.0, with 2.5 and 3.0 additionally
tested for some emotions.  The searches compared alternative layer locations
and both two- and three-layer configurations rather than assuming a common
layer set for all emotions.

Candidate configurations were first filtered by factual-preservation
constraints: mean token-level factual F1 had to be at least 0.85, and the mean
semantic factual score had to be at least 1.8.  Among feasible candidates, the
selected configuration maximized the mean improvement in target-emotion score
over the corresponding unsteered baseline.  Several emotion classes also had
held-out or independent confirmation runs.  However, the preserved development
artifacts do not establish one identical validation split and sample size for
all 11 classes; we therefore do not claim a single uniform validation protocol
across emotions.  The final configurations in
Table~\ref{tab:steering-layers} were fixed before the main evaluation.

The final steering artifact is constructed using the accepted training
memories only.
Its steering values average the emotion-neutral hidden-state differences of
the 6,927 accepted training memories; all 1,103 test memories are excluded from
value construction.

\subsection{NPTI}

NPTI acts on the intermediate FFN activation
$h_{\ell,j,t}=\operatorname{SiLU}((W_{\mathrm{gate}}x_{\ell,t})_j)$ of
Qwen2.5-7B-Instruct.  A neuron is counted as active at token $t$ exactly when
$h_{\ell,j,t}>0$; no positive-magnitude threshold beyond zero is used.  For
each emotion, activation frequencies are computed over all tokens in the
accepted emotion descriptions and their paired neutral controls:
\begin{equation}
\begin{aligned}
f^{\mathrm{emo}}_{\ell,j}
  &=\frac{1}{T_{\mathrm{emo}}}\sum_t
    \mathbb{I}[h^{\mathrm{emo}}_{\ell,j,t}>0],\\
f^{\mathrm{neu}}_{\ell,j}
  &=\frac{1}{T_{\mathrm{neu}}}\sum_t
    \mathbb{I}[h^{\mathrm{neu}}_{\ell,j,t}>0],\\
\delta_{\ell,j}&=f^{\mathrm{emo}}_{\ell,j}-f^{\mathrm{neu}}_{\ell,j}.
\end{aligned}
\end{equation}
Positive neurons satisfy $\delta_{\ell,j}>0.05$ and negative neurons satisfy
$\delta_{\ell,j}<-0.05$, independently at each of the 28 layers.  The formal
bank imposes no per-layer top-$k$ or other fixed count limit.

\begin{table}[t]
  \centering
  \normalsize
  \setlength{\tabcolsep}{6pt}
  \begin{tabular*}{\columnwidth}{@{\extracolsep{\fill}}lrr}
    \toprule
    \textbf{Emotion} & \textbf{Positive} & \textbf{Negative} \\
    \midrule
    admiration     & 38,692 & 84,563 \\
    amusement      & 42,196 & 88,041 \\
    confusion      & 45,201 & 73,497 \\
    curiosity      & 46,983 & 61,291 \\
    desire         & 52,317 & 81,486 \\
    disappointment & 42,663 & 79,087 \\
    embarrassment  & 48,637 & 79,182 \\
    excitement     & 44,594 & 86,566 \\
    joy            & 39,747 & 76,568 \\
    sadness        & 45,039 & 72,445 \\
    surprise       & 49,503 & 89,213 \\
    \bottomrule
  \end{tabular*}
  \caption{Numbers of selected positive and negative NPTI neurons, summed over
  all 28 transformer layers.  Selection is threshold-based rather than
  top-$k$.}
  \label{tab:npti-neuron-counts}
\end{table}

NPTI intervenes at the last-token position at every autoregressive decoding
step and across all 28 FFN layers.  For a selected positive neuron $j$ at layer
$\ell$, the intervention adds
\begin{equation}
  \Delta h_{\ell,j}=\gamma a^{95}_{\ell,j}\,
  \sigma\!\left(10\left(|\delta_{\ell,j}|-0.15\right)\right),
\end{equation}
where $a^{95}_{\ell,j}$ is the 95th percentile of the corresponding neutral
activation distribution and $\gamma=1.0$.
Selected negative neurons are clipped by
$h_{\ell,j}\leftarrow\min(h_{\ell,j},0)$.  Within a layer, all selected
positive neurons are updated and all selected negative neurons are clamped
simultaneously before the FFN down projection; the model then proceeds through
the 28 layers in normal Transformer order.

The value 0.05 is the neuron-selection threshold.  Threshold variants 0.10,
0.15, 0.20, and 0.25 were also materialized during development, but the
preserved artifacts do not record a uniform held-out selection rule establishing
0.05 as an automatically optimized value.  The separate constant 0.15 is the
fixed midpoint of the sigmoid strength weighting, and 10 is its slope; 0.15 is
not an additional neuron-selection threshold.  No separate validation record
for choosing this midpoint is preserved.

Neuron sets are constructed from the same 6,927 accepted training memories used
for steering-vector estimation.
The per-emotion sample counts sum to 6,927, and no test record contributes to
neuron selection.

\subsection{Checkpoints, Decoding, and Runtime}

The generator is Qwen/Qwen2.5-7B-Instruct, loaded in float16 from the local
ModelScope checkpoint.
The emotion encoder/evaluator is cirimus/modernbert-base-go-emotions, and the
NLI evaluator is Xenova/DeBERTa-v3-large-mnli-fever-anli-ling-wanli.

Generation uses the Qwen chat template with one query at a time and applies
greedy decoding with a maximum of 96 new tokens.
Sampling options, including temperature, top-$p$, top-$k$, and beam search, are
disabled.
Because decoding is deterministic, no generation sampling seed is required.
Retrieval-query encoding uses batch size 32 and maximum length 512.
ModernBERT cue encoding uses batch size 32 and maximum length 128, generation evaluation uses batch size 32 and maximum length 512, and NLI evaluation uses batch size 16 and maximum length 512.

The final evaluation pipeline used Python 3.11.7, PyTorch 2.9.0+cu128, CUDA 12.8, cuDNN 9.10.2, and Transformers 4.57.1.
Experiments were conducted on a server equipped with eight NVIDIA RTX A6000 GPUs, each providing approximately 48 GB of memory; individual jobs used subsets of the available GPUs as required by the workload.
The main prototype base seed is 20260922, with class-specific seed $20260922+r$.
Question generation uses seed 42; bootstrap seeds are 20260928 for the main results and 20260929 for the oracle decomposition.

\section{Qualitative Cases}
\label{app:cases}

Table~\ref{tab:cases} shows two Emotion Label examples selected by the largest paired change within a case type.
They illustrate the metrics and do not add statistical evidence.

\begin{table*}[t]
  \centering
  \footnotesize
  \setlength{\tabcolsep}{4pt}
  \begin{tabular}{p{1.3cm}p{1.8cm}p{1.5cm}p{0.9cm}p{3.5cm}p{5.1cm}}
    \hline
    \textbf{Condition} & \textbf{Injected} & \textbf{Predicted} & \textbf{Ent.} & \textbf{Event memory} & \textbf{Response (excerpt)} \\
    \hline
    \multicolumn{6}{p{14.6cm}}{Case 1 (Emotion-only retrieval). Target: admiration. Reference: ``[NAME], with his 8th appearance he's the most veteran player there.''} \\
    \hline
    Normal & confusion & curiosity & 0.018 & Ask about the relationship between FPTP and the referendum. & It sounds like you're referring to a specific event where someone named [NAME] is making his eighth appearance \ldots \\
    Oracle Emotion & admiration & admiration & 0.981 & Ask about the relationship between FPTP and the referendum. & It sounds like [NAME] has a significant presence and a strong track record at the event! \ldots \\
    Oracle Event & confusion & admiration & 0.999 & [NAME] is making his 8th appearance and is the most veteran player there. & Given that [NAME] is making his eighth appearance and is the most veteran player there \ldots \\
    Full Oracle & admiration & admiration & 0.999 & [NAME] is making his 8th appearance and is the most veteran player there. & That's a great opportunity to express your admiration and appreciation for [NAME]'s dedication \ldots \\
    \hline
    \multicolumn{6}{l}{Case 2 (Two-stage retrieval). Target: sadness. Reference: ``[NAME] wept.''} \\
    \hline
    Normal & disappointment & sadness & 0.000 & The situation is stranger than fiction, and [NAME] was going through something. & It sounds like [NAME] faced a challenging situation that became overwhelming, leading to a feeling of disappointment \ldots \\
    Oracle Emotion & sadness & neutral & 0.040 & The situation is stranger than fiction, and [NAME] was going through something. & I understand that [NAME] was going through a very challenging situation \ldots \\
    Oracle Event & disappointment & neutral & 1.000 & [NAME] wept. & When [NAME] became overwhelmed, they wept, likely due to a strong emotional response such as disappointment \ldots \\
    Full Oracle & sadness & neutral & 1.000 & [NAME] wept. & When [NAME] became overwhelmed, they wept. This suggests that [NAME] was experiencing a strong emotional response \ldots \\
    \hline
  \end{tabular}
  \caption{Two Emotion Label cases across the four oracle conditions.}
  \label{tab:cases}
\end{table*}

In Case 2 the Normal response is classified as sadness, which matches the target label, but it rests on an unrelated event and receives an entailment score near zero.
This case illustrates a response-level emotion match despite incorrect
episode retrieval and affect assignment.

\section{Protocol Audit}
\label{app:protocol}

Table~\ref{tab:audit} summarizes the protocol audit of the completed experiments.


\end{document}